\documentclass[12pt,draftcls,onecolumn]{IEEEtran}
\usepackage{subfigure}
\usepackage{graphicx,epsfig}
\usepackage{algorithm}
\usepackage{algorithmic}
\usepackage{amsmath}
\usepackage{multirow}
\usepackage{color}
\usepackage{bbding}

\ifCLASSINFOpdf
\else
\fi

\begin{document}

\title{\LARGE{Writer-Aware CNN for Parsimonious HMM-Based Offline Handwritten Chinese Text Recognition}}
\author{Zi-Rui~Wang,
        Jun~Du\Envelope,
        Jia-Ming~Wang
 \thanks{Zi-Rui Wang, Jun Du, and Jia-Ming Wang are with the National Engineering Laboratory for Speech and Language Information Processing, University of Science and Technology of China, Hefei, Anhui, P. R. China; email: cs211@mail.ustc.edu.cn, jundu@ustc.edu.cn, jmwang66@mail.ustc.edu.cn} }


\maketitle

\begin{abstract}
Recently, the hybrid convolutional neural network hidden Markov model (CNN-HMM) has been introduced for offline handwritten Chinese text recognition (HCTR) and has achieved state-of-the-art performance.
However, modeling each of the large vocabulary of Chinese characters with a uniform and fixed number of hidden states requires high memory and computational costs and makes the tens of thousands of HMM state classes confusing. Another key issue of CNN-HMM for HCTR is the diversified writing style, which leads to model strain and a significant performance decline for specific writers. To address these issues, we propose a writer-aware CNN based on parsimonious HMM (WCNN-PHMM). First, PHMM is designed using a data-driven state-tying algorithm to greatly reduce the total number of HMM states, which not only yields a compact CNN by state sharing of the same or similar radicals among different Chinese characters but also improves the recognition accuracy due to the more accurate modeling of tied states and the lower confusion among them. Second, WCNN integrates each convolutional layer with one adaptive layer fed by a writer-dependent vector, namely, the writer code, to extract the irrelevant variability in writer information to improve recognition performance. The parameters of writer-adaptive layers are jointly optimized with other network parameters in the training stage, while a multiple-pass decoding strategy is adopted to learn the writer code and generate recognition results. Validated on the ICDAR 2013 competition of CASIA-HWDB database, the more compact WCNN-PHMM of a 7360-class vocabulary can achieve a relative character error rate (CER) reduction of 16.6\% over the conventional CNN-HMM without considering language modeling. By adopting a powerful hybrid language model (N-gram language model and recurrent neural network language model), the CER of WCNN-PHMM is reduced to 3.17\%. Moreover, the state-tying results of PHMM explicitly show the information sharing among similar characters and the confusion reduction of tied state classes. Finally, we visualize the learned writer codes and demonstrate the strong relationship with the writing styles of different writers. To the best of our knowledge, WCNN-PHMM yields the best results on the ICDAR 2013 competition set, demonstrating its power when enlarging the size of the character vocabulary.

\end{abstract}


\IEEEpeerreviewmaketitle

\section{Introduction}

\IEEEPARstart{T}{he} robust recognition of handwritten text lines in an unconstrained writing style plays an important role in many applications, such as machine scoring, express sorting and document recognition. Specifically, handwritten Chinese text recognition (HCTR) has been intensively studied as a popular research topic for many years \cite{Fujisawa08,Liu18}. However, it remains a challenging problem due to the large vocabulary and the diversity of writing styles. Moreover, offline HCTR, which is the focus of this study, is more difficult than online HCTR {\cite{Jin17,LiuOn13}}, as the ink trajectory information is missing.

In general, the research efforts for offline HCTR can be divided into two categories: oversegmentation-based approaches and segmentation-free approaches. The former approaches \cite{Li10,Wang12,Wang16,Yi17} often build several modules by first including character oversegmentation, character classification, and modeling the linguistic and geometric contexts, and then incorporating them to calculate the score for path search. The recent work in \cite{Yi17}, with the neural network language model, adopted three different CNN models to replace the conventional character classifier, segmentation and geometric models to achieve the best performance of oversegmentation-based methods on the ICDAR 2013 competition dataset \cite{YinLiu13}. By contrast, segmentation-free approaches do not need to explicitly segment text lines. One early approach to text line modeling \cite{Su09} used the Gaussian mixture model hidden Markov model (GMM-HMM). Another recent approach \cite{Messina15} utilized multidimensional long short-term memory recurrent neural network (MDLSTM-RNN), which was inspired by well-verified LSTM-RNN approaches \cite{Graves09} for the recognition of handwritten western languages with a small set of character classes. The MDLSTM-RNN approach is quite flexible due to the connectionist temporal classification (CTC) technique \cite{Graves06}, which avoids explicit segmentation. In \cite{Hey2016}, the authors employed a CNN and an LSTM neural network under the HMM framework to obtain a significant improvement over the LSTM-HMM model. In \cite{Liu17ICDAR}, the authors used separable MDLSTM-RNN (SMDLSTM-RNN) with CTC loss, instead of the traditional LSTM-CTC method. More recently, the authors in \cite{JinCVPR19} proposed a novel aggregation cross-entropy loss for sequence recognition, which was shown to exhibit competitive performance for offline HCTR. In \cite{ZiRui18}, we verified that combining hybrid deep CNN-HMM (DCNN-HMM) with a powerful language model could achieve the best reported results of the segmentation-free approaches on the ICDAR 2013 competition dataset.

However, the impressive results reported in recently proposed oversegmentation-based and segmentation-free approaches \cite{Yi17}, \cite{JinCVPR19}, \cite{ZiRui18} highly depend on the use of strong language models (LMs) built with a large number of text corpora, which partially masks the weakness of character models and makes the comparison of character models unfair. Actually, the large vocabulary of Chinese characters and the diversified writing styles of text lines still limit the performance of deep learning methods based on character modeling. For example, in our DCNN-HMM work \cite{ZiRui18}, the number of output nodes in DCNN, i.e., the total state class number, was 19900 by modeling 3980 characters with a 5-state HMM for each. Obviously, a further increase of the vocabulary size could potentially lead to a data sparsity problem and high computation and memory costs, which makes the training of CNNs become difficult. Moreover, similar radicals among different Chinese characters should be shared by the same states to reduce ambiguity in the decoding stage.
Another key issue is that free-style writing usually causes a mismatch between the distributions of the training and testing datasets, which significantly degrades the recognition accuracy of certain writers.

To address these two main problems, we propose a novel writer-aware CNN based on parsimonious HMM (WCNN-PHMM). First, PHMM is designed using a data-driven state-tying algorithm to freely compress the total number of HMM states. The binary decision tree with a data-driven question set is adopted to represent one fixed-position HMM state of all character classes. In this way, it can not only yield a compact CNN by state sharing of the same or similar radicals among different Chinese characters but also improve the recognition accuracy due to the more accurate modeling of tied states and the lower confusion among them. Second, WCNN embeds one linear adaptive layer fed by a writer-dependent vector (namely, the writer code) into each convolutional layer, which extracts the irrelevant variability of writer information to improve recognition performance. In the training stage, all writer codes and the parameters of the adaptation layers are initialized randomly and then jointly optimized with other network parameters using the writer-specific data. In the recognition stage, with the initial recognition results from the first-pass decoding with the writer-independent CNN-PHMM model, an unsupervised adaptation is performed to generate the writer code for the subsequent decoding of WCNN-PHMM. Furthermore, in order to overcome the data sparseness problem of traditional N-gram LM (NLM) \cite{Katz1985}, similar to \cite{Yi17}, we introduce a recurrent neural network LM (RNNLM) \cite{Mikolov10} to form a hybrid LM (HLM).

The main contributions of this study can be summarized as follows:
\begin{itemize}
\item The new structure WCNN-PHMM is presented to tackle two key issues for offline HCTR: the large vocabulary and the diversity of writing styles.
\item A general adaptive training approach is proposed to integrate with any type of CNNs to create writer-aware models. To the best of our knowledge, this paper is the first study of writer adaptation for offline HCTR.
\item The fast and compact design of PHMM via state tying improves the recognition accuracy. More importantly, compared with other segmentation-free approaches, PHMM can yield even better recognition accuracy when enlarging the size of the character vocabulary by fully leveraging more training data and class information sharing.
\item The effectiveness of WCNN-PHMM is visually illustrated by the analyses of the state-tying results and the learned writer codes.
\item The proposed WCNN-PHMM demonstrates the best reported character error rate (CER) (8.42\%) for a 7360-class vocabulary on the ICDAR 2013 competition set without using language models. By adopting a powerful HLM, the CER of WCNN-PHMM can be further reduced to 3.17\%.
\end{itemize}

The remainder of this paper is organized as follows. Section~\ref{sec:rw} introduces related work. Section~\ref{sec:sys_over} gives an overview of the proposed framework. Section~\ref{sec:WCNN-PHMM} elaborates on the details of WCNN-PHMM. Section~\ref{sec:exp} reports the experimental results and analyses. Finally, Section~\ref{sec:con} concludes.

\section{Related Work}
\label{sec:rw}
In this section, we describe related work, including the basic principles for mainstream approaches of offline HCTR, model compression and writer adaptation.

\subsection{Basic principles for offline HCTR}
Offline HCTR can be formulated as the Bayesian decision problem:
\begin{eqnarray} \label{basic_principle}
\begin{aligned}
\displaystyle \hat{\mathbf{C}} &= \arg\max_{\mathbf{C}} p(\mathbf{C} | \mathbf{X})  \\
                               &= \arg\max_{\mathbf{C}}p(\mathbf{X} | \mathbf{C}) p(\mathbf{C})
\end{aligned}
\end{eqnarray}
where $\mathbf{X}$ is the feature sequence of a given text line image and $\mathbf{C}=\{C_1, C_2, ... , C_n\}$ is the underlying $n$-character sequence. In oversegmentation-based approaches \cite{Wang12}, the posterior probability $p(\mathbf{C} | \mathbf{X})$ can be computed by searching the optimal segmentation path and the corresponding posterior probability of the character sequence by combining the character classifier, the segmentation model and the geometric/language model. Regarding segmentation-free approaches, the CTC-based and HMM-based approaches are two mainstream frameworks. In the CTC-based approach \cite{Liu17ICDAR}, a special character blank class and a defined many-to-one mapping function are introduced to directly compute $p(\mathbf{C} | \mathbf{X})$ with the forward-backward algorithm \cite{Graves06}. For the HMM-based approach \cite{ZiRui18},  $p(\mathbf{C} | \mathbf{X})$ can be reformulated as the conditional probability $p(\mathbf{X} | \mathbf{C})$ and the prior probability $p(\mathbf{C})$. More details will be provided in Section~\ref{sec:sys_over}.

\subsection{Model compression}
The state tying can be regarded as belonging to a more general field, i.e., model compression \cite{Bucilua2006}. With the emergence of deep learning \cite{Hin2015}, many studies have focused on building compact and fast CNNs for practicability.
Regarding the reduction in the number of parameters and the computation complexity of convolutional layers, research efforts can be divided roughly into low-rank decomposition \cite{Zhang16}, pruning \cite{He17}, quantization \cite{leng2017} and compact network design \cite{zhang_2017}. Aside from these methods, a key issue with CNN-HMM-based offline HCTR \cite{ZiRui18} is the large vocabulary problem, which leads to tens of thousands of output nodes (corresponding to HMM states) in CNN architecture. This heavy overhead in the output layer of the CNN not only requires high memory and computation costs but also yields more confusion among state classes and CNN training difficulties. To handle this problem, inspired by the early work in speech recognition \cite{Young94,HTK}, we introduce state tying via decision trees to freely compress the output layer of the CNN model. Considering the particularity of HCTR and the difficulty of defining an effective question set for the Chinese language, in our previous work \cite{Wenchao}, we successfully invented a data-driven state-tying approach for a huge set of HMMs representing Chinese characters and achieved promising recognition performance. It should be noted that, if we simply reduce the state number for each character, the recognition accuracy will decline dramatically due to the lack of resolution for text line modeling \cite{ZiRui18}.

\begin{figure}
\centering
\includegraphics[width=3.3in]{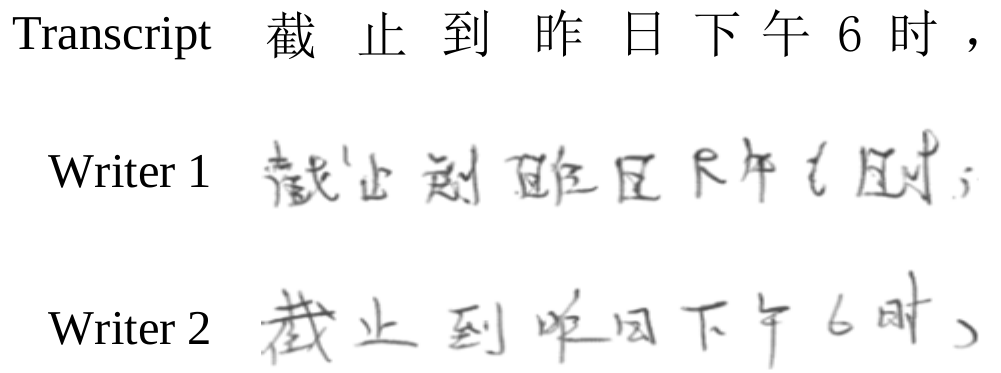}
\caption{Handwritten examples of different writers with the same transcript.}
\label{different_writer}
\end{figure}

\subsection{Writer adaptation}
Writer adaptation is similar to other topics, such as transfer learning \cite{TransLearn} and speaker adaptation \cite{SpeakerAdap}, where the distribution of test data is different from that of training data \cite{ZhangLiu13}. In offline HCTR, as shown in Fig.~\ref{different_writer}, the writing styles could be quite different, which makes the recognition accuracy of unseen writers unpredictable. In comparison to handwritten Chinese character recognition (HCCR), aside from the morphological variations within characters, writing orientation and ligatures make HCTR much more challenging. In general, there are two mainstream methodologies to achieve writer adaptation. The one type is to adopt writer-specific data to guide writer-independent classifier toward the new distribution of the particular writer, the other is to extract writer-independent features for classifier. More specifically, this process might be supervised, semisupervised or unsupervised, depending on whether the adaptation writer-specific data are labeled. Usually, unsupervised adaptation needs to reuse the test data. Besides, it depends on adequate writer data. In some applications such as the machine scoring of essays \cite{bridgeman2012comparison}, the recognition rate is the most important factor to be considered and there are enough specific writer data available to adopt adaptation techniques for improving the recognition rate. Moreover, the research on writer adaptation could be divided into feature-space and model-space approaches based on the part on which the adaptation parameters are working \cite{Du13}. To the best of our knowledge, for Chinese handwriting recognition, almost all efforts of writer adaptation focus on the HCCR task. One such method uses a linear feature transformation to adapt the writing styles via discriminative linear regression (DLR) \cite{Du14,Du15}, which is verified to be effective when incorporated with a prototype-based classifier and an NN-based classifier. Another representative method introduces style transfer mapping (STM) \cite{ZhangLiu13} for learning a linear transformation to project writer-specific data onto a style-free space. As a flexible adaptation method, STM can work on the outputs of both fully connected layers \cite{HongMing16,XuYao2016} and convolutional layers \cite{Yang18}. A recent study \cite{Ian2016} uses adversarial learning \cite{Ian2016} to transform writer-dependent features into writer-independent features under the guidance of printed data. However, there are very few studies for the writer adaptation of the more challenging HCTR problem. Inspired by \cite{JH13,Xue14}, in \cite{ZiRui16} we propose an unsupervised writer adaptation strategy for DNN-HMM-based HCTR.

This study is comprehensively extended from our previous conference papers \cite{Wenchao,ZiRui16} with the following new contributions: 1) the proposed PHMM is introduced with more technical details and verified for a more promising CNN-HMM, rather than the DNN-HMM in \cite{Wenchao}; 2) we present a novel unsupervised adaptation strategy with writer codes and adaptation layers to guide the convolutional layers in CNN-HMM, rather than using the fully connected layers in DNN-HMM \cite{ZiRui16}; 3) WCNN-PHMM perfectly combines the two techniques to yield a compact and high-performance model; 4) instead of the NLM, the HLM is used to further improve performance; and 5) all experiments are redesigned to verify the effectiveness of WCNN-PHMM, and detailed analyses are described to give the readers a deep understanding of our approach.

\section{System Overview}
\label{sec:sys_over}

\begin{figure}
\centering
\includegraphics[width=5in]{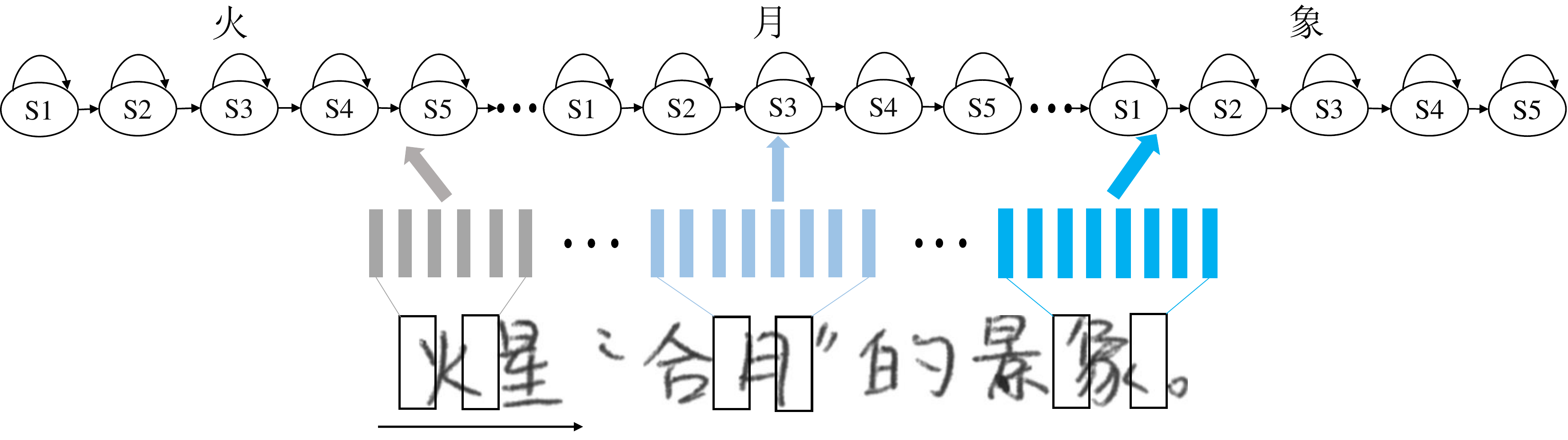}
\caption{Illustration of text line modeled by cascading character HMMs.}
\label{hmm}
\end{figure}

Our system follows the basic HMM framework \cite{ZiRui18} in which the handwritten text line is modeled by a series of cascading HMMs, each representing one character, as illustrated in Fig.~\ref{hmm}. The mathematic principle of HMM can be represented by rewriting the formula $p(\mathbf{X} | \mathbf{C}) p(\mathbf{C})$ in Eq.~(\ref{basic_principle}):
\begin{eqnarray} \label{beys}
 p(\mathbf{X} | \mathbf{C}) p(\mathbf{C})  &=&\sum\limits_{S} \left[ \pi(s_0) \prod_{t=1}^{T} a_{s_{t-1}s_t}\prod_{t=0}^{T} p(\mathbf{x}_t | s_t)\right] \prod_{i=1}^n p(C_i|C_{i-1}, C_{i-2}, ... , C_1) \label{eq:bayesian1} \\
&=&\sum\limits_{S} \left[ \pi(s_0) \prod_{t=1}^{T} a_{s_{t-1}s_t}\prod_{t=0}^{T} \frac{p(s_t|\mathbf{x}_t)p(\mathbf{x}_t)}{p(s_t)} \right]  \prod_{i=1}^n p(C_i|C_{i-1}, C_{i-2}, ... , C_1) \label{eq:bayesian2}
\end{eqnarray}
where $\mathbf{X}=\{\mathbf{x}_0, \mathbf{x}_1, \mathbf{x}_2, ... ,\mathbf{x}_T\}$ is a $(T+1)$-frame observation sequence of one text line image. $p(\mathbf{X} |
\mathbf{C})$, which can be called the character model, is the
conditional probability of $\mathbf{X}$ given $\mathbf{C}$
corresponding to a sequence of HMMs with the corresponding hidden state sequence $S=\{s_0, s_1, s_2, ... ,s_T\}$. Each HMM with a set of states
represents one character class. With HMMs, the $p(\mathbf{X} |
\mathbf{C})$ can be decomposed in the frame level: $\pi(s_0)$ is the initial state probability, $a_{s_{t-1}s_t}$ is the state transition probability from frame $t-1$ to $t$, $p(\mathbf{x}_t | s_t)$ is the output probability of $\mathbf{x}_t$ given $s_t$,
$p(s_t)$ is the prior probability of state $s_t$ estimated from the training set,
$p(s_t|\mathbf{x}_t)$ is the posterior probability of state $s_t$ given $\mathbf{x}_t$,
 and $p(\mathbf{x}_t)$ is independent of the character sequence. As mentioned in \cite{ZiRui18}, GMM can be used to calculate $p(\mathbf{x}_t | s_t)$ in Eq.~(\ref{eq:bayesian1}) for the GMM-HMM system, while DNN/CNN can be adopted to compute $p(s_t|\mathbf{x}_t)$ in Eq.~(\ref{eq:bayesian2}) for the DNN-HMM/CNN-HMM system.

Meanwhile, $p(\mathbf{C})$, namely the language model, is the probability of an $n$-character sequence $\mathbf{C}=\{C_1,C_2,...,C_n\}$ and can be decomposed as $\prod_{i=1}^n p(C_i|C_{i-1}, C_{i-2}, ... , C_1)$. However, as the number of these values $V^{i}$ for even a moderate vocabulary size $V$ is too large to be accurately estimated. The so-called N-gram LM can not realistically depend on all $i-1$ conditioning histories $C_1, C_2, ... ,C_{i-1}$ to compute the term $p(C_i|C_{i-1}, C_{i-2}, ... , C_1)$. Obviously, a higher order $N$ leads to a more powerful language model which can significantly improve the recognition accuracy. In this work, the SRILM toolkit \cite{Stolcke02} is employed to generate a 5-gram LM. To further enhance the ability of the LM, we linearly interpolate a standard NLM with an RNNLM to form an HLM.

In the training stage, we first build the conventional GMM-HMM system as in \cite{ZiRui18}. Then, the state-tying GMM-HMM system (GMM-PHMM) can be generated using the proposed decision-tree algorithm to greatly reduce the total number of states, i.e., the dimension of the CNN output layer. Meanwhile, state-level forced-alignment is conducted to obtain frame-level labels for the subsequent CNN cross-entropy training. After the conventional CNN is trained, a series of adaptation layers with the writer codes as the input are appended in parallel to form the WCNN. With writer-specific training data, the writer codes and the parameters of the adaptation layers for WCNN are jointly optimized.

In the testing stage, with the initial recognition results from the first-pass decoding using CNN-PHMM, the codes of unknown writers are learned from random initialization via WCNN for the second-pass decoding. This process could be iteratively conducted for multipass decoding to refine the recognition results and the writer codes.

\section{WCNN-PHMM}
\label{sec:WCNN-PHMM}

\begin{figure*}
\centering
\includegraphics[width=6in]{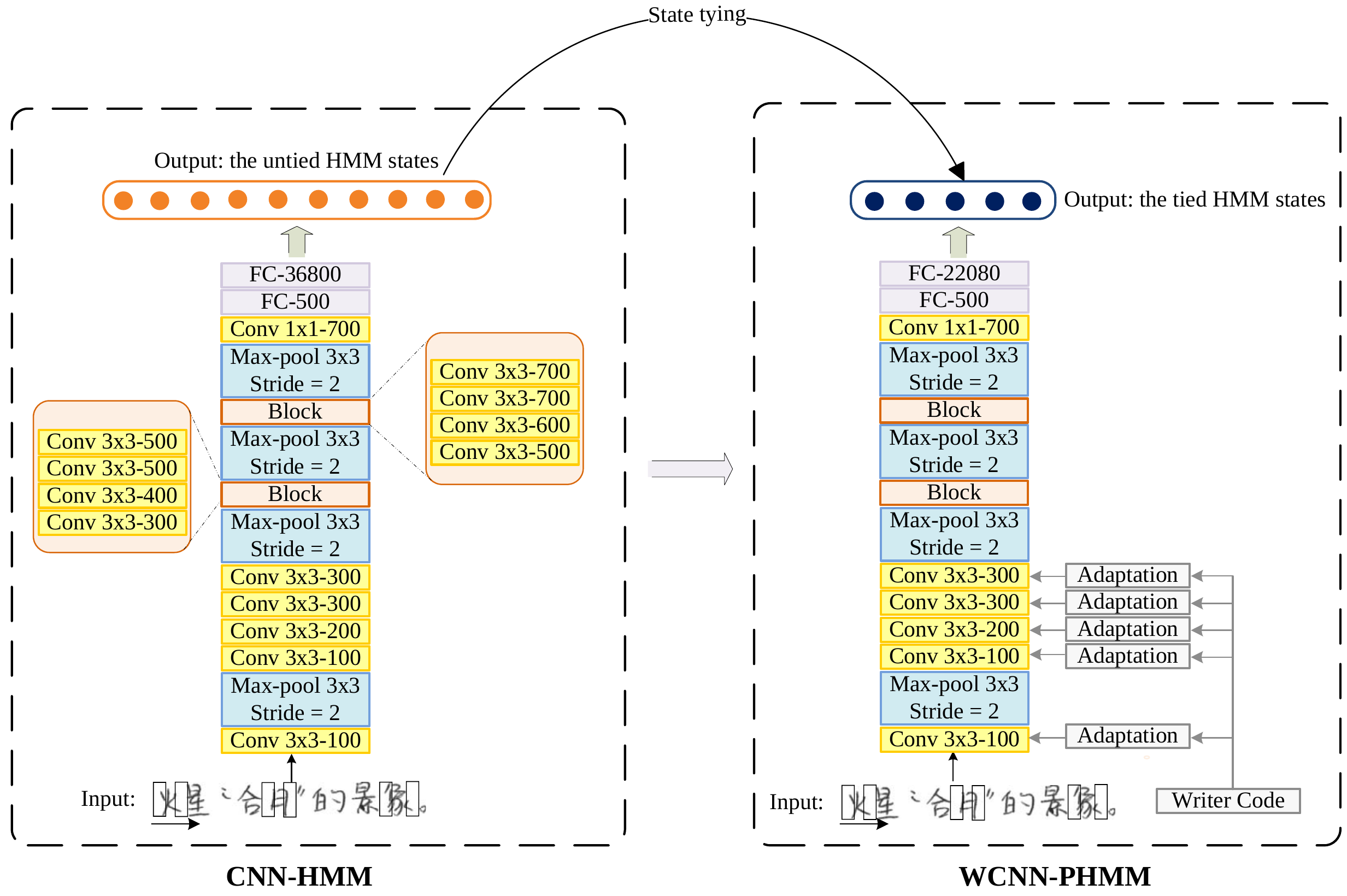}
\caption{{Comparison between the conventional CNN-HMM and the proposed WCNN-PHMM.}}
\label{development_WCNN_PHMM}
\end{figure*}

Fig.~\ref{development_WCNN_PHMM} illustrates two main innovations of our proposed WCNN-PHMM architecture over the conventional CNN-HMM in \cite{ZiRui18}, namely, the compact design of the output layer and writer-aware convolutional layers. In the following subsections, we elaborate three basic components of WCNN-PHMM: convolutional neural network, state tying for PHMM, and writer code-based adaptive training for WCNN. In order to help readers understand clearly, in Table~\ref{AD}, we first describe acronyms that are frequently used in this paper. For example, according to Table~\ref{AD}, the system WCNN-PHMM means characters are modeled by the PHMM where the WCNN is used to compute the posterior probabilities of tied-states.

\begin{table}
\caption{Acronym Description}
\centering \label{AD}
\begin{tabular}{|c|c|}
\hline
Acronym             & Description  \\
\hline
CNN        & Convolutional neural network               \\
\hline
WCNN      & Writer-aware convolutional neural network \\
\hline
TCNN    &   Tied-state convolutional neural netwotk \\
\hline
HMM        & Hidden Markov model               \\
\hline
PHMM        & Parsimonious hidden Markov model              \\
\hline
CER      & Character error rate \\
\hline
\end{tabular}
\end{table}

\begin{figure}
\centering
\includegraphics[width=4in]{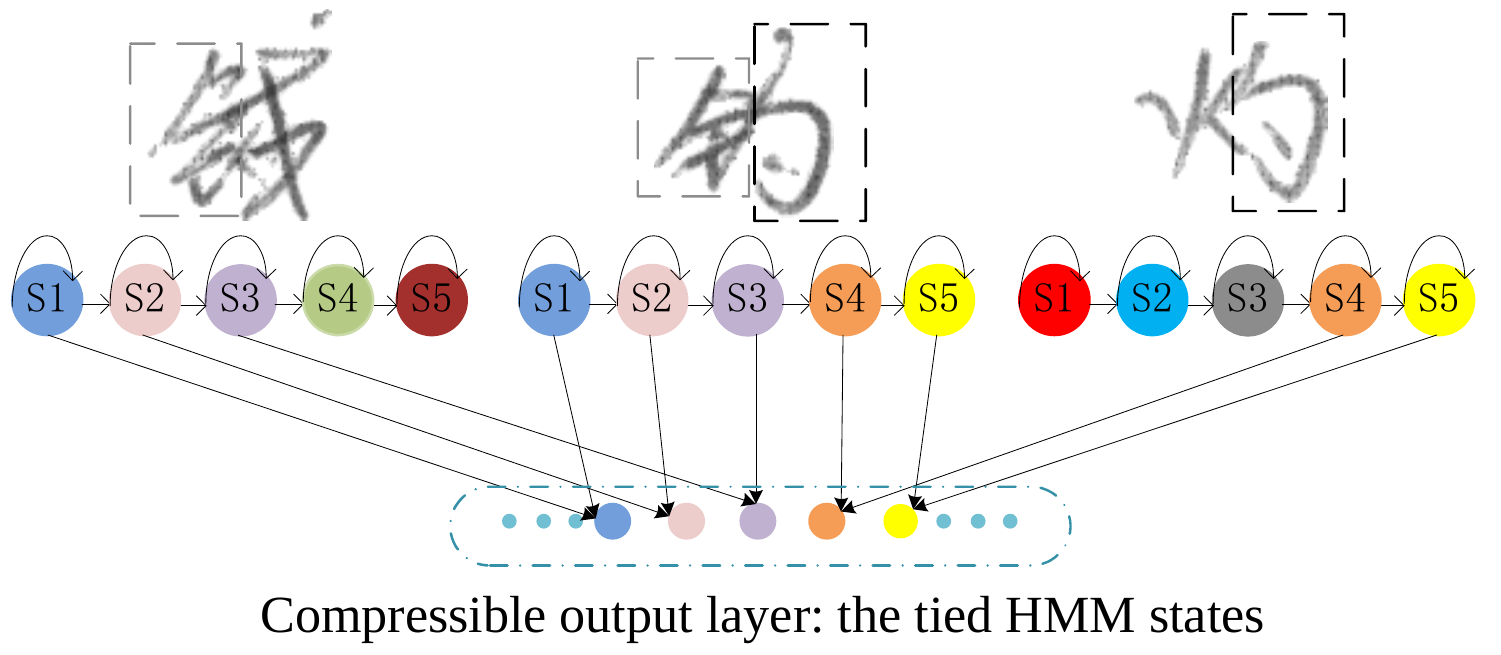}
\caption{Illustration of tied state design for CNN output layer.}
\label{PHMM}
\end{figure}

\subsection{Convolutional neural network}
As shown in Fig.~\ref{development_WCNN_PHMM}, CNN \cite{Yann98} successively consists of stacked convolutional layers (Conv) optionally followed by spatial pooling, one or more fully connected layer (FC) and a softmax layer. For the convolutional and pooling layers, each layer is a three-dimensional tensor organized by a set of planes called feature maps, while the fully connected layer and the softmax layer are the same as those in the conventional DNN. Inspired by the locally sensitive, orientation-selective neurons in the visual system of cats \cite{Hubel1962}, each unit in a feature map is constrained to connect a local region in the previous layer, which is called the local receptive field. Two contiguous local receptive fields are usually $s$ pixels (referred as stride) shifted in a certain direction. Usually, all units in the same feature map of a convolutional layer share a set of weights, each computing a dot product between its weights and the local receptive field in the previous layer and then followed by batch normalization (BN) \cite{BN} and a nonlinear activation function. Meanwhile, the units in a pooling layer perform a spatial average or max operation for their local receptive field to reduce spatial resolution and noise interference. Accordingly, the key information for identifying the pattern is retained.
 We formalize operations in a convolutional layer as:
\begin{eqnarray}
\label{convlay}
{\boldsymbol{O}_{i,j,k}}
= f (\text{BN} (\sum\limits_{m,n,l} {{\boldsymbol{I}_{(i - 1) \times s + m,(j - 1) \times s + n,l}}{\boldsymbol{W}_{m,n,k,l}}}+\boldsymbol{B}_k ))
\end{eqnarray}
where ${\boldsymbol{I}_{i,j,k}}$ is the value of the input unit in feature map $k$ at row $i$ and column $j$ while ${\boldsymbol{O}_{i,j,k}}$ corresponds to the output unit, $\boldsymbol{W}_{m,n,k,l}$ is the connection weight between a unit in feature map $k$ of the output and a unit in channel $l$ of the input, with an offset of $m$ rows and $n$ columns between the output unit and the input unit. $\boldsymbol{B}_k$ is the $k$-th value of bias vector $\boldsymbol{B}$ for all units in the feature map $k$. BN is used to handle the change of the distribution in each layer by simply normalizing the input of layers \cite{BN}, which can yield an obvious improvement in the HCTR task \cite{ZiRui18}. $f$ is a nonlinear function, i.e., ReLU \cite{Krizhevsky2012}, used in this study.
\subsection{State tying for PHMM}
Fig.~\ref{PHMM} illustrates the main motivation of our proposed algorithm to tie HMM states, namely, fully utilizing the partial similarities of characters (e.g., radicals). State tying is completed using a binary decision tree in which the question for each node of the tree is automatically generated by a data-driven algorithm. If each character is represented by a 5-state HMM, then 5 trees are built, with each representing one positioned HMM state to cluster all character classes. Suppose $\mathbf{S}$ is the set of HMM states in one nonleaf node of a tree and $L({\mathbf{S}})$ is the log-likelihood of $\mathbf{S}$ generating the training dataset with $F$ frames. Then, by the attached question $q$, which is selected from an automatically generated question set, this node with $\mathbf{S}$ is split into two children nodes, namely, a left node with a subset $\mathbf{S}_{\text{l}}$ and a right node with a subset $\mathbf{S}_{\text{r}}$, to maximize the log-likelihood increase with respect to $q$ in the current node:
\begin{eqnarray}
\label{loglin}
\Delta L = L({\mathbf{S}_{\text{l}}(q)}) + L({\mathbf{S}_{\text{r}}(q)}) - L(\mathbf{S})
\end{eqnarray}
where $L({\mathbf{S}})$, $L({\mathbf{S}_{\text{l}}(q)})$ and $L({\mathbf{S}_{\text{r}}(q)})$, are log-likelihoods of the state set in the current node, its left node and its right node, respectively. Based on the assumptions that all tied states in $\mathbf{S}$ share a common mean $\boldsymbol{\mu}(\mathbf{S})$ and variance $\boldsymbol{\Sigma}(\mathbf{S})$, and the tying states does not change the frame/state alignment, a reasonable approximation of $L(\mathbf{S})$ via Gaussian output distribution $\mathcal{N}$ is given by:
\begin{align}
\label{logl}
 L(\mathbf{S}) &= \sum\limits_{f=1}^F \sum\limits_{s \in \mathbf{S}} {\gamma _s}({\boldsymbol{o}_f})\ln \mathcal{N}({\boldsymbol{o}_f};\boldsymbol{\mu}(\mathbf{S}),\boldsymbol{\Sigma}(\mathbf{S}))  \nonumber \\
&= -\frac{1}{2}\sum\limits_{f=1}^F \sum\limits_{s \in \mathbf{S}} {\gamma _s}({\boldsymbol{o}_f}) [D\ln(2\pi) + \ln |\boldsymbol{\Sigma(\mathbf{S})}| + D^2_M(\boldsymbol{o}_f)]
\end{align}
where $D_M(\boldsymbol{o}_f)$ is the Mahalanobis distance:
\begin{eqnarray}
D_M(\boldsymbol{o}_f) = \sqrt{(\boldsymbol{o}_f - \boldsymbol{\mu}(\mathbf{S}) )^{\top} (\boldsymbol{\Sigma}(\mathbf{S}))^{-1} (\boldsymbol{o}_f - \boldsymbol{\mu}(\mathbf{S}))} .
\end{eqnarray}
In Eq.~(\ref{logl}), $\gamma_s(\boldsymbol{o}_f)$ is the posterior probability of the $D$-dimensional feature vector $\boldsymbol{o}_f$ at the $f$-th frame that is generated by state $s$. $\boldsymbol{\mu}(\mathbf{S})$ and $\boldsymbol{\Sigma(\mathbf{S})}$ can be estimated as:
\begin{eqnarray}
\label{globalmean} {\mu}(\mathbf{S}) &=& \frac{\sum\limits_{f=1}^F \sum\limits_{s \in \mathbf{S}} {\gamma _s}({\boldsymbol{o}_f}) \boldsymbol{o}_f}{\sum\limits_{f=1}^F \sum\limits_{s \in \mathbf{S}} {\gamma _s}({\boldsymbol{o}_f})}  \\
\label{globalvar}\boldsymbol{\Sigma(\mathbf{S})} &=& \frac{\sum\limits_{f=1}^F \sum\limits_{s \in \mathbf{S}} {\gamma _s}({\boldsymbol{o}_f}) (\boldsymbol{o}_f - \boldsymbol{\mu}(\mathbf{S}) ) (\boldsymbol{o}_f - \boldsymbol{\mu}(\mathbf{S}) )^{\top}}{\sum\limits_{f=1}^F \sum\limits_{s \in \mathbf{S}} {\gamma _s}({\boldsymbol{o}_f})}.
\end{eqnarray}

Using Eq.~(\ref{globalvar}), we can have the following derivation for the last item in Eq.~(\ref{logl}):
\begin{align}
&\sum\limits_{f=1}^F \sum\limits_{s \in \mathbf{S}} {\gamma _s}({\boldsymbol{o}_f}) D^2_M(\boldsymbol{o}_f) \nonumber\\
=& \sum\limits_{f=1}^F \sum\limits_{s \in \mathbf{S}} {\gamma _s}({\boldsymbol{o}_f}) \text{Tr}\{(\boldsymbol{o}_f - \boldsymbol{\mu}(\mathbf{S}) )^{\top} (\boldsymbol{\Sigma}(\mathbf{S}))^{-1} (\boldsymbol{o}_f - \boldsymbol{\mu}(\mathbf{S}))\} \nonumber\\
=& \sum\limits_{f=1}^F \sum\limits_{s \in \mathbf{S}} {\gamma _s}({\boldsymbol{o}_f}) \text{Tr}\{ (\boldsymbol{\Sigma}(\mathbf{S}))^{-1} (\boldsymbol{o}_f - \boldsymbol{\mu}(\mathbf{S})) (\boldsymbol{o}_f - \boldsymbol{\mu}(\mathbf{S}) )^{\top}\} \nonumber\\
=& \text{Tr}\{ (\boldsymbol{\Sigma}(\mathbf{S}))^{-1}  \sum\limits_{f=1}^F \sum\limits_{s \in \mathbf{S}} {\gamma _s}({\boldsymbol{o}_f})   (\boldsymbol{o}_f - \boldsymbol{\mu}(\mathbf{S})) (\boldsymbol{o}_f - \boldsymbol{\mu}(\mathbf{S}) )^{\top} \}  \nonumber\\
=& \text{Tr}\{ (\boldsymbol{\Sigma}(\mathbf{S}))^{-1}  \boldsymbol{\Sigma}(\mathbf{S}) \}\sum\limits_{f=1}^F \sum\limits_{s \in \mathbf{S}} {\gamma _s}({\boldsymbol{o}_f})  =  D\sum\limits_{f=1}^F \sum\limits_{s \in \mathbf{S}} {\gamma _s}({\boldsymbol{o}_f})
\end{align}
where $\text{Tr}\{\cdot\}$ denotes the trace of a square matrix. If we further define the notation:
\begin{eqnarray}
\label{globalocc} {\gamma}(\mathbf{S}) &=& \sum\limits_{f=1}^F \sum\limits_{s \in \mathbf{S}} {\gamma _s}({\boldsymbol{o}_f})
\end{eqnarray}
Then, Eq.~(\ref{logl}) can be rewritten as:
\begin{eqnarray}
\label{loglfinal}
 L(\mathbf{S}) &=& -\frac{1}{2}{\gamma}(\mathbf{S}) [\ln |\boldsymbol{\Sigma(\mathbf{S})}| + D + D\ln(2\pi)]
\end{eqnarray}
Thus, the log-likelihood $L(\mathbf{S})$ depends only on the pooled state occupancy ${\gamma}(\mathbf{S})$ and the pooled state variance $\boldsymbol{\Sigma(\mathbf{S})}$. Both could be calculated from the saved parameters of state occupancy counts, means, and variances for all HMM states during the preceding Baum-Welch re-estimation.

\begin{figure}
\centering
\includegraphics[width=6in]{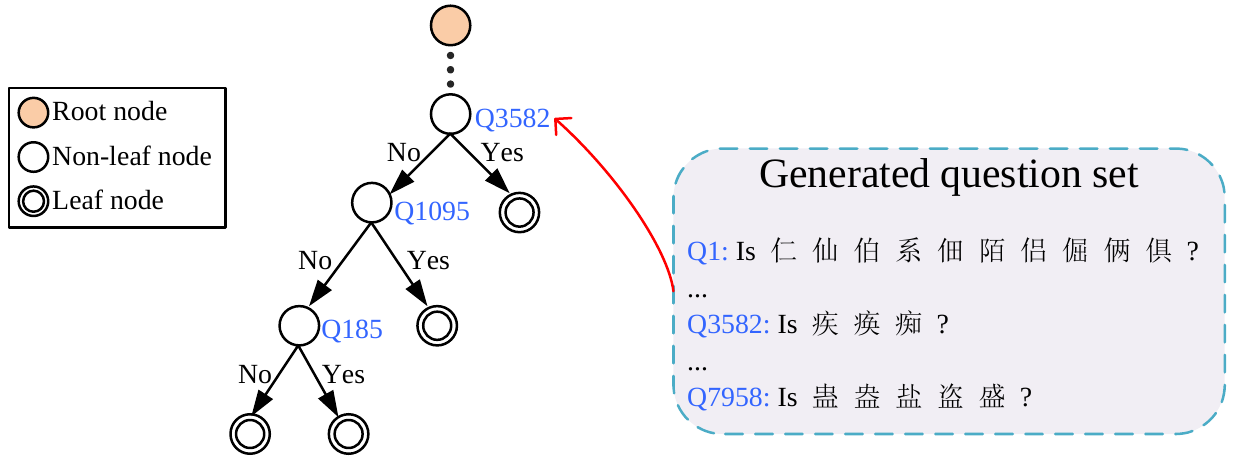}
\caption{Fraction of a generated tree for the first state of a 5-state HMM.}
\label{Tree}
\end{figure}

Initially, all corresponding states are placed in the root node of a tree. Then, the above algorithm is conducted in a top-down manner to build this binary tree until reaching to a fixed threshold. Finally, a merge operation of leaf nodes is conducted using a minimum priority queue in a bottom-up manner by computing the log-likelihood decrease to reach the target tied-state number.

To generate the question set, all feature frames of characters are placed in the root node of a binary decision tree and then a $k$-means ($k=2$) algorithm is used to find an optimal partition, which aims to maximize the log-likelihood of frames under the assumption of a single Gaussian distribution. This procedure is conducted in a top-down manner until each node only contains one character class. One question of a nonleaf node can be obtained from all reachable leaves of this node. All questions form our question set for the state tying. There are 5 trees in total, as each character is modeled by a 5-state HMM. In Fig.~\ref{Tree}, a fraction of a generated tree for the first state is illustrated.

In Table~\ref{Diff_HCTR_SR}, we summarize the differences of state tying between HCTR and speech recognition (SR). First, the original signal in HCTR is two-dimension image and the signal is one-dimension speech in SR. Second, the motivation of state tying in HCTR is to overcome the difficulty of training and decoding in CNN-HMM due to many similar radicals among tens of thousands of characters while the state tying in SR is introduced for the data sparseness problem of tri-phone. Third, considering the ways of modeling in HCTR, we only tie the states of characters being in the same position to capture similar radicals more accurately. For SR, the state tying is usually conducted on the states of tri-phones with the same central phone. Finally, for HCTR, the question set used in state tying totally depends on the character based features while the question set in SR can be predefined artificially according to pronunciation characteristics.

\begin{table}
\caption{The differences of state tying in HCTR and SR}
\centering \label{Diff_HCTR_SR}
\begin{tabular}{|c|c|c|}
\hline
             & HCTR & SR    \\
\hline
Original Signal & Two dimension  & One dimension \\
\hline
Object         & The states of characters being in the same position      &   The states of tri-phones with the same central phone \\
\hline
Motivation      & Existing similar radicals among characters &  Data sparseness problem of tri-phone \\
\hline
Categories    &   Tens of thousands & Hundreds  \\
\hline
Question Set        & Data driven     &    Date driven or Artificial rules      \\
\hline
\end{tabular}
\end{table}

\subsection{Adaptive training for WCNN based on writer code}

As shown in Fig.~\ref{development_WCNN_PHMM}, the conventional CNN used for offline HCTR does not explicitly incorporate the writer information in both training and testing stages. However, the writing style could play an essential role in the final CER as an irrelevant variability to recognize the character class. Accordingly, a learnable vector (writer code) is introduced to represent the writer style of each writer. If we consider the CNN architecture to integrate both feature extraction and classifier implicitly, then the proposed ingenious design of WCNN in Fig.~\ref{development_WCNN_PHMM} seems like a joint feature and model adaptive training strategy.


\begin{figure}
\centering
\includegraphics[width=5.5in]{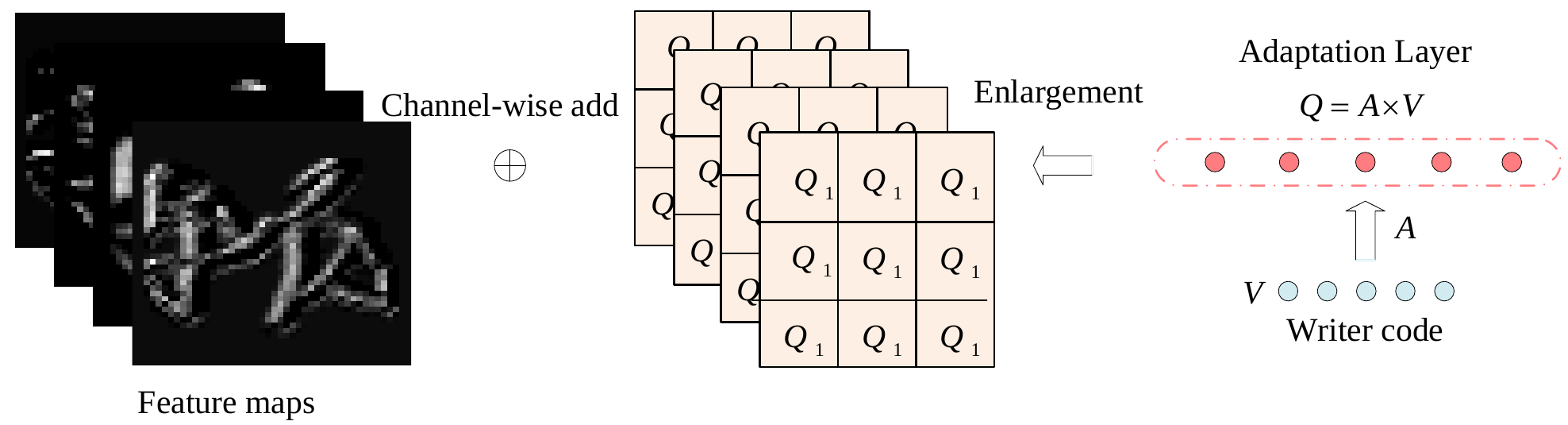}
\caption{Illustration of convolutional layer with writer code in WCNN.}
\label{WCNN}
\end{figure}

To guide the CNN with writer information, two key components, i.e, writer codes and adaptation layers, are randomly initialized and can be optimized using the back-propagation algorithm. The code of the $r$-th writer is a $G$-dimensional vector $\boldsymbol{V}^{r}$ directly connected with all adaptation layers. The $p$-th adaptation layer can be represented by a $K \times G$ matrix $\boldsymbol{A}^{p}$. The writer code is fed into the adaptation layer and transformed into a new vector $\boldsymbol{Q}^{r,p}$:

\begin{eqnarray}
\label{writer_adap}
\boldsymbol{Q}^{r,p}=\boldsymbol{A}^{p}\boldsymbol{V}^{r}.
\end{eqnarray}
With the writer information $\boldsymbol{Q}$, the corresponding $p$-th convolutional layer of WCNN can be reformulated as:
\begin{eqnarray}
\label{conv_ada}
{\boldsymbol{O}_{i,j,k}^{r,p}} = f(\text{BN}( \boldsymbol{M}_{i,j,k}^{p} + {\boldsymbol{Q}_k^{r,p}}))
\end{eqnarray}
where
\begin{eqnarray}
\label{ori_conv}
{\boldsymbol{M}_{i,j,k}^{p}} = \sum\limits_{l,m,n} {{\boldsymbol{I}_{(i - 1) \times s + m,(j - 1) \times s + n,l}^{p}}{\boldsymbol{W}_{m,n,k,l}^{p}}}  + {\boldsymbol{B}_k^{p}}.
\end{eqnarray}
In Eqs.~(\ref{conv_ada}-\ref{ori_conv}), ${\boldsymbol{I}_{i,j,k}^p}$, ${\boldsymbol{O}_{i,j,k}^p}$, $\boldsymbol{W}_{m,n,k,l}^p$, and $\boldsymbol{B}_k^p$ are the corresponding items like in Eq.~(\ref{convlay}) for the $p$-th convolutional layer. The writer information ${\boldsymbol{Q}_k^{r,p}}$,  which is the $k$-th value of bias vector $\boldsymbol{Q}^{r,p}$, is newly added as a bias to build writer-aware convolutional layers. The key innovation of the WCNN architecture is illustrated in Fig.~\ref{WCNN}.

Suppose we use $P$ adaptation layers with the parameter set $\boldsymbol{A}=\{\boldsymbol{A}^{p}|p=1,...,P\}$. In the training stage, a well-trained CNN-HMM or CNN-PHMM system is first used to initialize WCNN with the writer-independent parameter set $\boldsymbol{\Lambda}$. Assume we have $R$ writers in the training dataset with the corresponding writer code set $\boldsymbol{V}=\{\boldsymbol{V}^{r}|r=1,...,R\}$. Then, the cross-entropy criterion is minimized with respect to writer-aware parameter set $\{\boldsymbol{A},\boldsymbol{V}\}$ in WCNN:
\begin{eqnarray}
\label{opt_adapara}
E(\boldsymbol{\boldsymbol{A},\boldsymbol{V}}) = - \sum\limits_{t=1}^{N_B} \log p(s_t|\boldsymbol{X}_t,\boldsymbol{\Lambda},\boldsymbol{A},\boldsymbol{V})
\end{eqnarray}
where the WCNN output $p(s_t|\boldsymbol{X}_t,\boldsymbol{\Lambda},\boldsymbol{A},\boldsymbol{V})$ is the posterior probability of the reference state $s_t$ given the input image $\boldsymbol{X}_t$ within the sliding window. $N_B$ is the minibatch size using stochastic gradient decent algorithm. In our implementation, we process the text lines one by one. Thus, $N_B$ equals the number of frames of each text line. Please note that, for each frame $\boldsymbol{X}_t$, the input parallel writer code vector is selected from $\boldsymbol{V}$ with the writer-aware information. With the random initialization, we jointly update $\{\boldsymbol{A},\boldsymbol{V}\}$ using backpropagation and SGD:
\begin{eqnarray}
\label{backprop_adapara}
\boldsymbol{A}^{p} \leftarrow \boldsymbol{A}^{p} - \varepsilon^{\text{tr}} \frac{\partial E(\boldsymbol{\boldsymbol{A},\boldsymbol{V}})}{\partial \boldsymbol{A}^{p}} \nonumber \\
\boldsymbol{V}^{r} \leftarrow \boldsymbol{V}^{r} - \varepsilon^{\text{tr}} \frac{\partial E(\boldsymbol{\boldsymbol{A},\boldsymbol{V}})}{\partial \boldsymbol{V}^{r}}
\end{eqnarray}
where $\varepsilon^{\text{tr}}$ is the step size in the training stage, which is initially set to 0.001 and decreased by a factor of 0.8 after updating with 5 million frames. We summarize the training procedure of WCNN in Algorithm~\ref{alg:algorithm_adap_train}.

\begin{algorithm}[htb]
\caption{The training procedure of WCNN.}
\label{alg:algorithm_adap_train}
\begin{algorithmic}[1]
\REQUIRE ~~\\
The writer-independent parameter set $\boldsymbol{\Lambda}$ is generated using conventional CNN-HMM/CNN-PHMM systems; \\
Randomly initialize the writer-aware parameter set $\{\boldsymbol{A},\boldsymbol{V}\}$;\\
Prepare the minibatch level training dataset with the state label and writer information in each frame,\\
\STATE Randomly select one minibatch and set the input writer code of each frame using writer information and $\boldsymbol{V}$.
\STATE Calculate all required derivatives using backpropagation.
\STATE Update the adaptation layer parameters and writer codes $\{\boldsymbol{A},\boldsymbol{V}\}$ using Eq.~(\ref{backprop_adapara}).
\STATE Go to step 1 until the convergence condition is satisfied.
\ENSURE The parameter set of WCNN  $\{\boldsymbol{\Lambda},\boldsymbol{A},\boldsymbol{V}\}$
\end{algorithmic}
\end{algorithm}

\begin{algorithm}[htb]
\caption{ The adaptation/recognition procedure of WCNN.}
\label{alg:algorithm_adap_te}
\begin{algorithmic}[1]
\REQUIRE ~~\\
Prepare the WCNN parameter set $\{\boldsymbol{\Lambda},\boldsymbol{A}\}$; \\
Prepare the minibatch level dataset of an unknown writer;\\
Randomly initialize the corresponding writer code $\boldsymbol{V}^{\text{U}}$,\\
\STATE Generate the state labels via first-pass decoding using $\boldsymbol{\Lambda}$.\
\STATE Perform the adaptation to refine $\boldsymbol{V}^{\text{U}}$ using Eq.~(\ref{backprop_adapara_test}).
\STATE Conduct decoding using $\{\boldsymbol{\Lambda},\boldsymbol{A},\boldsymbol{V}^{\text{U}}\}$ of WCNN.
\STATE Go to step 2 for alternative adaptation and recognition until a specified number of multipass decoding is reached.
\ENSURE The writer code $\boldsymbol{V}^{\text{U}}$ and recognition results
\end{algorithmic}
\end{algorithm}

In the recognition stage, for the data of an unknown writer, a multipass decoding is conducted. In the first-pass decoding, we use only CNN-HMM/CNN-PHMM with the parameter set $\boldsymbol{\Lambda}$ to generate the recognition results that are adopted as the state labels for updating the writer code vector of this unknown writer in the next pass. In the second pass, we perform the adaptation by minimizing the cross-entropy criterion with respect to the writer code $\boldsymbol{V}^{\text{U}}$:
\begin{eqnarray}
\label{opt_adapara_test}
E'(\boldsymbol{V}^{\text{U}}) = - \sum\limits_{t=1}^{N'_B} \log p(s^{\text{U}}_t|\boldsymbol{X}^{\text{U}}_t,\boldsymbol{\Lambda},\boldsymbol{A},\boldsymbol{V}^{\text{U}}).
\end{eqnarray}
Similar to Eq.~(\ref{opt_adapara}), $\boldsymbol{X}^{\text{U}}_t$ is the $t$-th input frame of an unknown writer, while $s^{\text{U}}_t$ is its corresponding state label from the first-pass recognition. The batch size $N'_B$ refers to the number of frames of each text line. Please note that we do not use $\boldsymbol{V}^{\text{U}}$ from the training stage and randomly initialize the code $\boldsymbol{V}^{\text{U}}$ of the unknown writer. Accordingly, we can update $\boldsymbol{V}^{\text{U}}$ as:
\begin{eqnarray}
\label{backprop_adapara_test}
\boldsymbol{V}^{\text{U}} \leftarrow \boldsymbol{V}^{\text{U}} - \varepsilon^{\text{ts}} \frac{\partial E'(\boldsymbol{V}^{\text{U}})}{\partial \boldsymbol{V}^{\text{U}}}
\end{eqnarray}
where $\varepsilon^{\text{ts}}$ is the step size in the testing stage, which is set to 0.001. Then, we conduct a second-pass decoding using $\{\boldsymbol{\Lambda},\boldsymbol{A},\boldsymbol{V}^{\text{U}}\}$ of WCNN. This adaptation and recognition processes could be alternatively and iteratively conducted until a specified number of multipass decoding is reached. We summarize the adaptation/recognition procedure of WCNN in Algorithm~\ref{alg:algorithm_adap_te}.

\subsection{Hybrid language model}
The HLM is linear interpolation of a traditional NLM and an RNNLM. Considering all calculations in Eq.~(\ref{beys}) are performed in the logarithmic domain, the HLM is represented as:
\begin{eqnarray}
\log {p_{\text{HLM}}}(\mathbf{C}) = \omega\log {p_{\text{NLM}}}(\mathbf{C}) + (1-\omega) \log {p_{\text{RNNLM}}}(\mathbf{C})
\end{eqnarray}
where the ${p_{\text{NLM}}}(\mathbf{C})$ means the probability of an $n$-character sequence $\mathbf{C}=\{C_1,C_2,...,C_n\}$ is computed based on NLM while the value of ${p_{\text{RNNLM}}}(\mathbf{C})$ is obtained from RNNLM. $\omega$ is a hyperparameter to adjust the ratio between NLM and RNNLM. In the RNNLM, a simple RNN with three layers including input layer, hidden layer and output layer is used. At time step $i$, the input vectors consist of a 1-of-$V$ coding $\boldsymbol{R}_i$ that represents the previous word $C_{i-1}$, and the previous hidden layer output $\boldsymbol{H}_{i-1}$. The output of the hidden layer is computed as:
\begin{eqnarray}
\boldsymbol{H}_{i}=f(\boldsymbol{W}_{H,V}\boldsymbol{R}_i+\boldsymbol{W}_{H,H}\boldsymbol{H}_{i-1})
\end{eqnarray}
where $\boldsymbol{W}_{H,V}$ and $\boldsymbol{W}_{H,H}$ are learnable matrices of size $H\times V$ and $H \times H$, respectively. The activation function $f$ is sigmoid. In the output layer, using the history information $\boldsymbol{H}_{i}$, the probabilities of the predicted characters at time step $i$ are estimated:
\begin{eqnarray}
\boldsymbol{P}_{i}=g(\boldsymbol{W}_{V,H}\boldsymbol{H}_{i})
\end{eqnarray}
$g$ is the softmax function and $\boldsymbol{W}_{V,H}$ is a $V \times H$ learnable matrix. Naturally, for a predicted character $C_i$ at time step $i$, we have the following equation:
\begin{eqnarray}
  p_\text{RNNLM}(C_i|C_{i-1}, C_{i-2}, ... , C_1) =\boldsymbol{P}_{i}(C_i).
\end{eqnarray}
Finally,
\begin{eqnarray}
{p_{\text{RNNLM}}}(\mathbf{C}) = \prod_{i=1}^n p_\text{RNNLM}(C_i|C_{i-1}, C_{i-2}, ... , C_1) = \prod_{i=1}^n \boldsymbol{P}_{i}(C_i).
\end{eqnarray}
In this work, the dimension of the hidden layer is set to 300, the $\omega$ is 0.5 and the weights \{$\boldsymbol{W}_{H,V}, \boldsymbol{W}_{H,H}, \boldsymbol{W}_{V,H}$\} in the RNNLM are optimized by using the truncated BPTT \cite{Mikolov11}.

\section{Experiments}
\label{sec:exp}
We designed a set of experiments to validate and explain the effectiveness of the proposed method for offline HCTR.
All experiments were implemented with Kaldi \cite{Kaldi} and Pytorch \cite{torch} toolkits using NVIDIA GeForce GTX 1080Ti GPUs. Additionally, we plan to release our source codes in the near future.

\subsection{Dataset and metrics}
\begin{table}
\caption{The information of the CASIA-HWDB databases.}
\centering \label{info}
\begin{tabular}{|c|c|c|c|c|c|}
\hline
\#             & Class          & Writer        & Text Line    & Character Sample  \\
\hline
HWDB1.0        & 3,837          & 420           & -            & 1,592,978      \\
\hline
HWDB1.1        & 3,834          & 300           & -            & 1,145,074      \\
\hline
HWDB2.0        & 1,222          & 419           & 20,495       & 540,468       \\
\hline
HWDB2.1        & 2,310          & 300           & 17,292       & 429,926       \\
\hline
HWDB2.2        & 1,331          & 300           & 14,443       & 383,153       \\
\hline
\end{tabular}
\end{table}

We conducted the experiments on a widely used database for HCTR released by the Institute of Automation of Chinese Academy of Sciences (CASIA) \cite{Liu11-2,Liu13}. To train the character models, both offline isolated handwritten Chinese character datasets (HWDB1.0 and HWDB1.1) and the training sets of offline handwritten Chinese text datasets (HWDB2.0, HWDB2.1, and HWDB2.2) were used. The detailed information, including the number of classes, writers, lines, and characters for each dataset, are shown in Table~\ref{info}. In total, 3,980 classes (Chinese characters, symbols, garbage) were formed with 4,091,599 samples. To train the language model, the training sets of offline handwritten Chinese text of HWDB2.0-2.2 and the news data downloaded from Internet are used. All the news data have been checked to exclude the text of the test set. The whole corpus contains approximately ten million characters.
The ICDAR 2013 competition set with 60 writers unseen to the training dataset was adopted as the evaluation set \cite{YinLiu13}. The CER was computed as:
\begin{equation}
\text{CER} = \frac{{{N_\text{s}} + {N_\text{i}} + {N_\text{d}}}}{N}
\end{equation}
where $N$ is total number of character samples in the evaluation set. $N_\text{s}$, $N_\text{i}$ and $N_\text{d}$ denote the number of substitution errors, insertion errors and deletion errors, respectively. Firstly, to focus on character modeling, we did not use additional language models.

\subsection{Experiments on state tying of PHMM}
\subsubsection{Comparison between CNN-HMM and CNN-PHMM}
We first compared CNN-HMM with CNN-PHMM according to the best configuration in our previous work \cite{ZiRui18}, i.e., there were 16 weight layers (14 Conv and 2 FC layers) and the number of channels increased from 100 to 700. The image patch of each frame was passed through a stack of 3$\times$3 convolutional layers. After the last max pooling layer, a 1$\times$1 convolutional layer was used to increase the nonlinearity of the net without more computation and memory than the other larger receptive fields. All convolutional layers were followed by the ReLU and the stride was 1, while the stride of all max pooling layers was 2 with a 3$\times$3 window. The BN operation was equipped for the outputs before nonlinearity in every convolutional layer. The minibatch size was 1,000, the momentum was 0.9 and the weight decay was 0.0001. The learning rate was initially set to 0.01 and decreased by 0.92 after every 4,000 batches.
Three epochs were conducted. All other parameters, such as frame length, frame shift, feature extraction for GMM-HMM, and parameters of GMM-HMM, were the same as those used in \cite{ZiRui18}.

For CNN-HMM, we list the results of different settings of states per HMM in Table~\ref{DiffStates_CER}. The observation consistent with \cite{ZiRui18} was that the CER increased greatly from 5 states to 1 state due to the lack of adequate resolution. Notably, the number of output nodes of CNN was 3,980$\times$5 (19,900) for 5-state HMM, while the number of output nodes was 3,980 for 1-state HMM, which means that, the more states for each character, the more challenging it is to train CNN. Based on the optimal settings of the 5-state CNN-HMM system, we conducted the state-tying algorithm of our PHMM to reduce the average number of states per HMM. Interestingly, the performance of CNN-PHMM could improve when the average number of states equaled 3 or 4; however, if we kept reducing this number to 2 or 1, the performance declined. These observations implied that there was a tradeoff between the model resolution and the parameter redundancy. Moreover, the CER of CNN-PHMM was much lower than the CER of CNN-HMM for the same average state number, which indicated that CNN-PHMM achieved more reasonable state assignment among all character HMMs than CNN-HMM. Another advantage of CNN-PHMM is its more compact CNN output layer, which helps compress the CNN and accelerate the decoding process, as shown in Table~\ref{DiffStates_MC_AD}. Finally, for CNN-PHMM, an average 3 states was used as the default for the subsequent experiments, which not only achieved a much lower CER than the best configured CNN-HMM with 5 states but also yielded a much smaller model size and a faster decoding speed.

\begin{table}
\caption{CER (\%) comparison between CNN-HMM and CNN-PHMM based on different settings of average states per HMM.}
\centering \label{DiffStates_CER}
\begin{tabular}{|c|c|c|c|c|c|}
\hline
\# of states per HMM               &  5      &  4          & 3       & 2   &  1  \\
\hline
CNN-HMM                                &  10.02      &  10.11        &    10.77      &    11.71  & 13.85  \\
\hline
CNN-PHMM        & 10.02  &      9.44      &     9.54        &  9.91   & 11.61    \\
\hline
\end{tabular}
\end{table}

\begin{table}
\caption{Practical issue comparison of different settings of average states per HMM for the corresponding CNN-PHMM system in Table~\ref{DiffStates_CER}. $N_{\text{M}}$ and $N_{\text{T}}$ represent the model size and run-time latency, respectively, which are normalized by those of CNN-HMM system with 5 states per HMM.}
\centering \label{DiffStates_MC_AD}
\begin{tabular}{|c|c|c|c|c|c|}
\hline
\# of states per HMM               &  5      &  4          & 3         & 2      &  1  \\
\hline
$N_{\text{M}}$             &  1      &  0.96       & 0.87      &  0.83  & 0.77  \\
\hline
$N_{\text{T}}$             &  1      &  0.91      &  0.72      &  0.63  & 0.57    \\
\hline
\end{tabular}
\end{table}

\subsubsection{Analysis of state tying}

\begin{figure}
\centering
\hspace{-0.05in}
\includegraphics[width=3.3in]{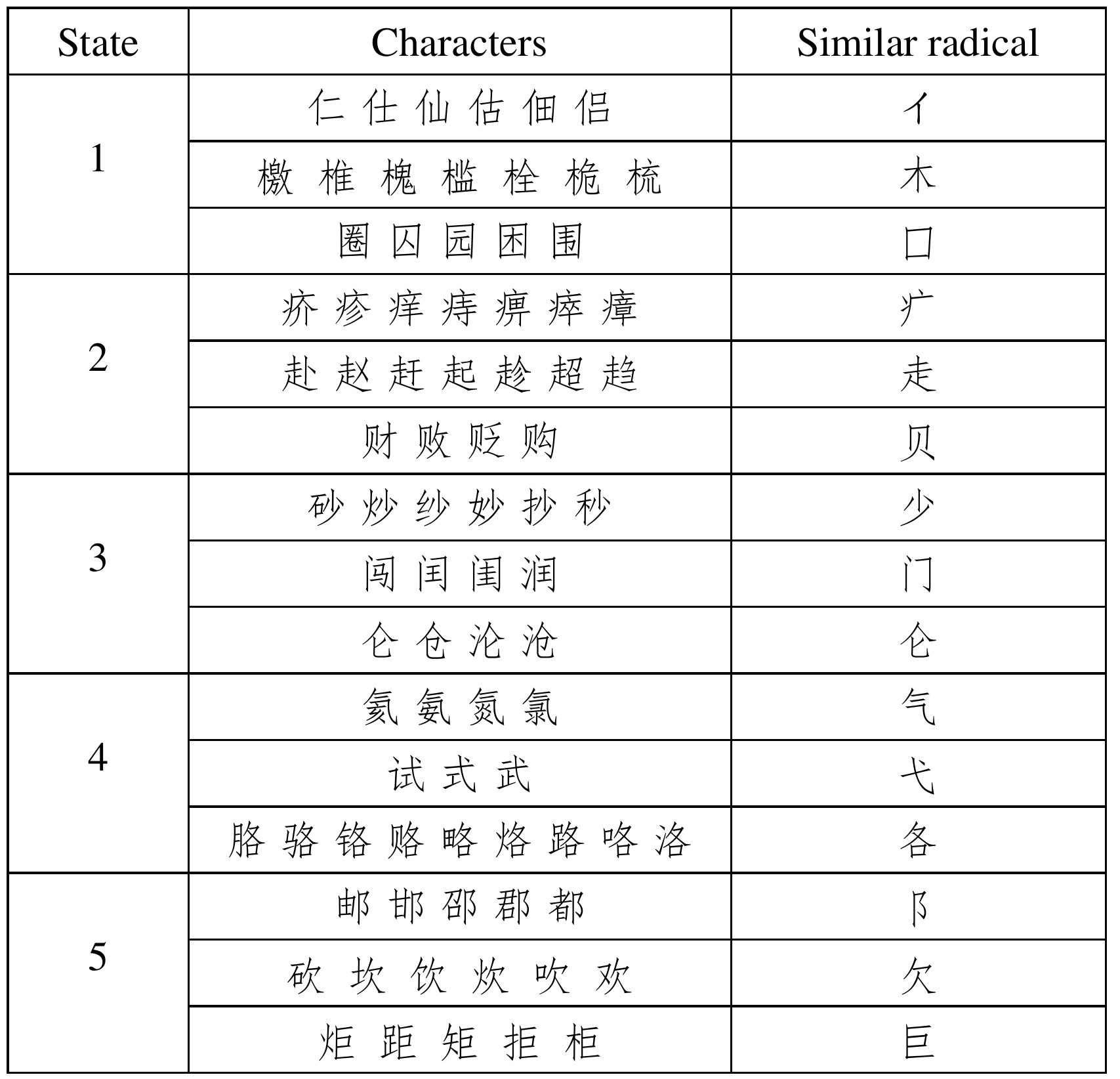}
\caption{Examples of tied Chinese characters with similar radicals.}
\label{tied_results}
\end{figure}

\begin{figure}
\centering
\hspace{0.3in}
\includegraphics[width=3.3in]{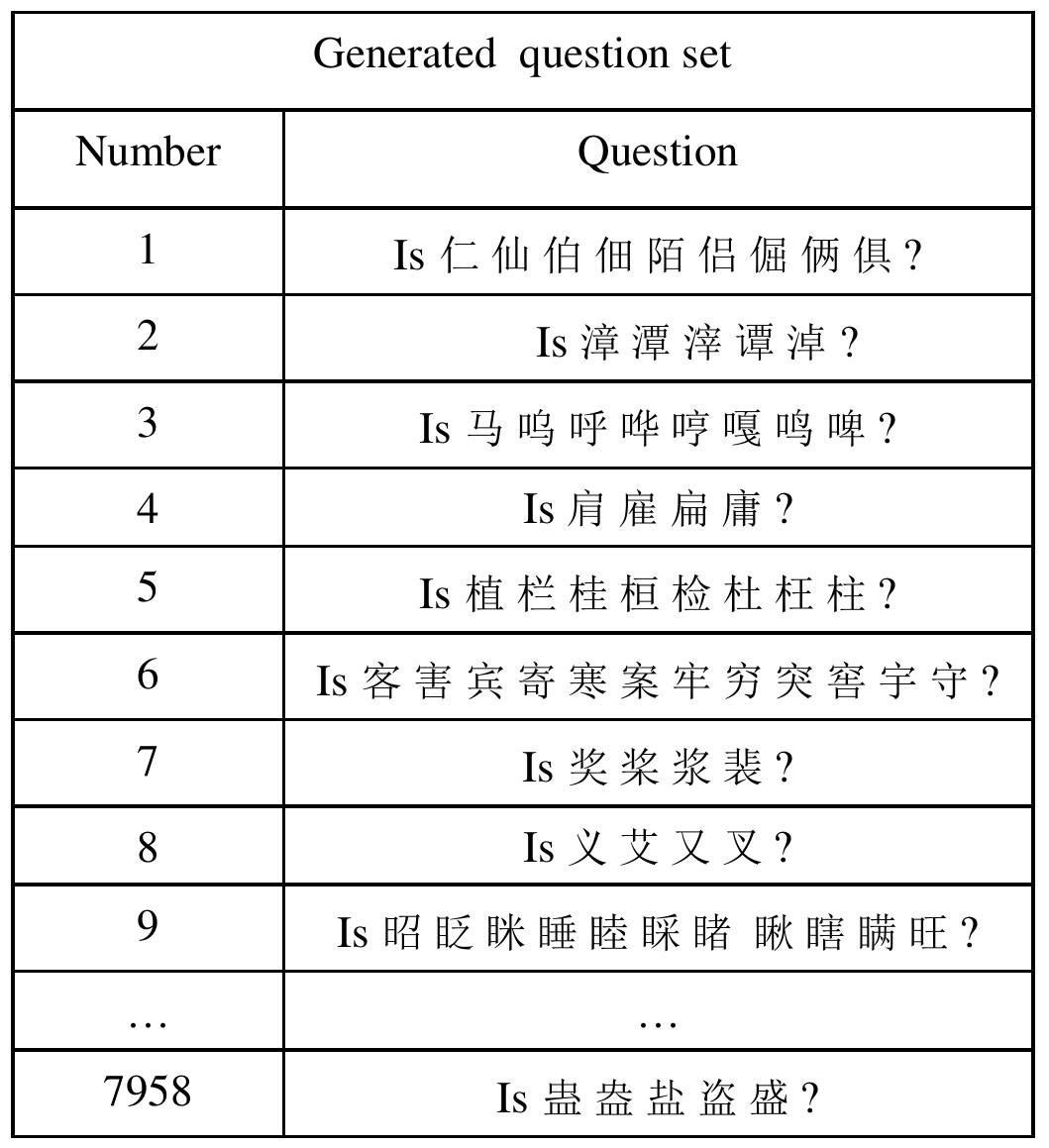}
\caption{Partial results of generated question set for tree-based state tying.}
\label{question_set}
\end{figure}

In Fig.~\ref{tied_results}, we list representative examples of tied Chinese characters from positioned states 1 to 5 in our CNN-PHMM system. It was quite intuitive and reasonable that most of the tied Chinese characters shared the same or similar radicals although the state-tying process was purely data driven with diversified writing styles. This result could explain why there was a large amount of parameter redundancy in the conventional untied CNN-HMM model. We also give partial results of the data-driven question set in Fig.~\ref{question_set}. In total, there were 7,938 questions generated. It could be observed that the related characters in one question were similar, which demonstrated the effectiveness of the $k$-means clustering algorithm. Overall, the proposed state-tying method has two advantages. First, because the total number of states corresponds to the size of the CNN output layer, having fewer categories will make CNN training easier and speed up the recognizer. Second, reducing parameter redundancy can potentially increase the number of training samples for the tied states from different characters.

\begin{figure}
\centering
\hspace{0.3in}
\includegraphics[width=3.4in]{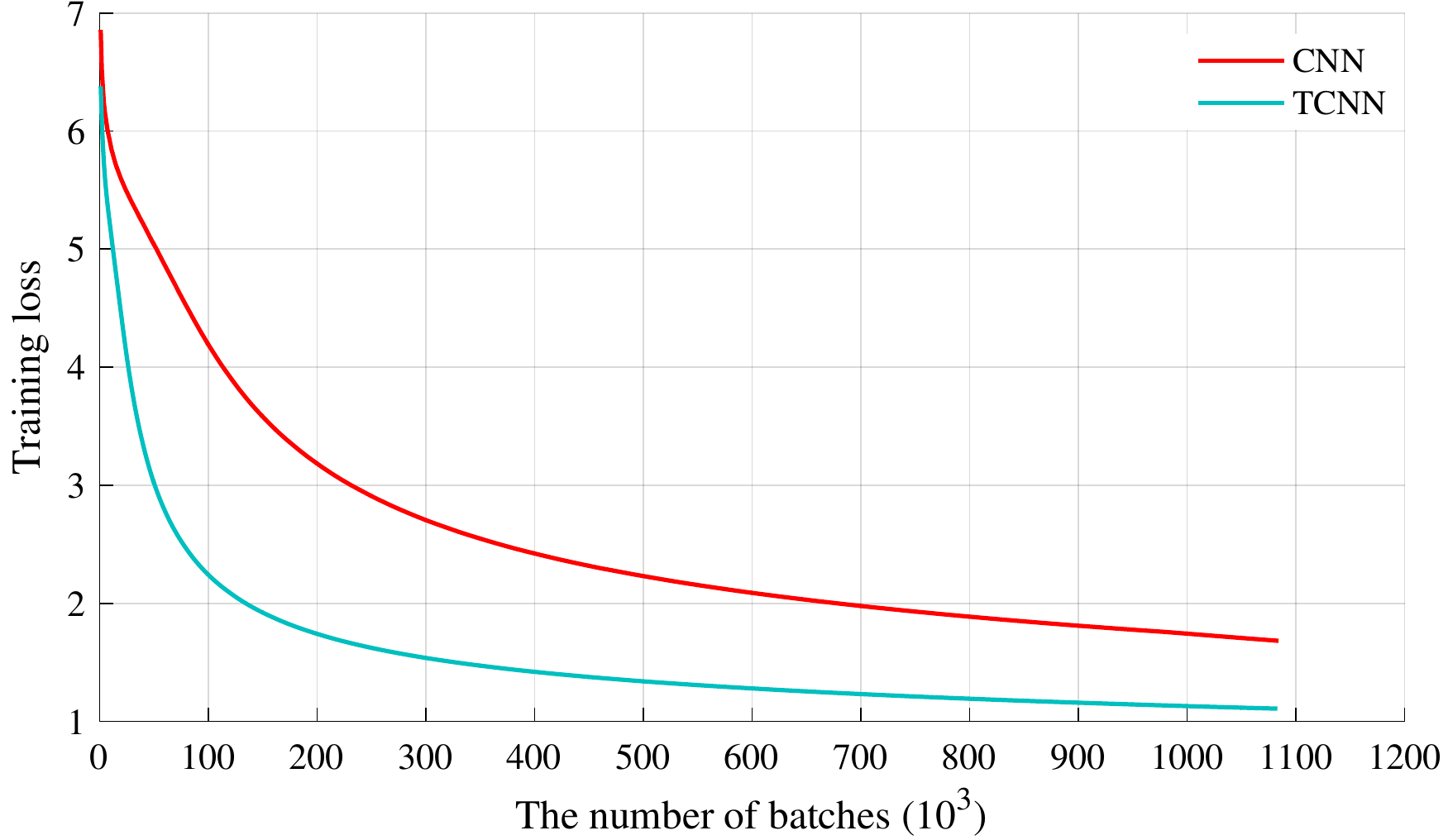}
\caption{Training loss comparison between CNN and TCNN.}
\label{training_loss}
\end{figure}

For further analysis, we draw the learning curves during training for conventional CNN and tied-state CNN (TCNN) in Fig.~\ref{training_loss}. Obviously, the learning curve of TCNN was always below that of CNN. More interestingly, the gap between the two curves significantly increased in the beginning stage and then decreased to a relatively stable value as an increasing amount of training data was used. We believe that the compact design of the CNN output layer not only made the CNN model easier to train and more effective to classify but also fully utilized the training data by state tying.

\subsection{Experiments on writer adaptive training for WCNN}

\subsubsection{The configuration of WCNN}
\label{conf_wcnn}
As shown in Fig.~\ref{PHMM}, there are two key factors for writer-adaptive training: the number of adaptation layers $P$ and the dimension of writer code $G$. The increase in the number of adaptation layers linking to the convolutional layers goes from input layer to output layer. Table~\ref{DiffLinks} compares different settings of adaptation layer number $P$ and writer code dimension $G$ in WCNN-PHMM. $P$=0 denotes the CNN-PHMM system without writer adaptive training. Please note that second-pass decoding was adopted as a default for WCNN-PHMM. When the writer code dimension was fixed as 200, the CER decreased from 9.54\% to 8.96\% with $P$ increasing from 0 to 5. The performance was saturated when more than 5 adaptation layers were used due to the limited adaptation data. Another interesting observation is that the performance of WCNN-PHMM was not sensitive to writer code dimension, with a good tradeoff of $G$=200. Thus, we use the configuration of $P$=5 and $G$=200 in the following experiments.

\begin{figure}
\centering
\hspace{-0.32in}
\includegraphics[width=4.5in]{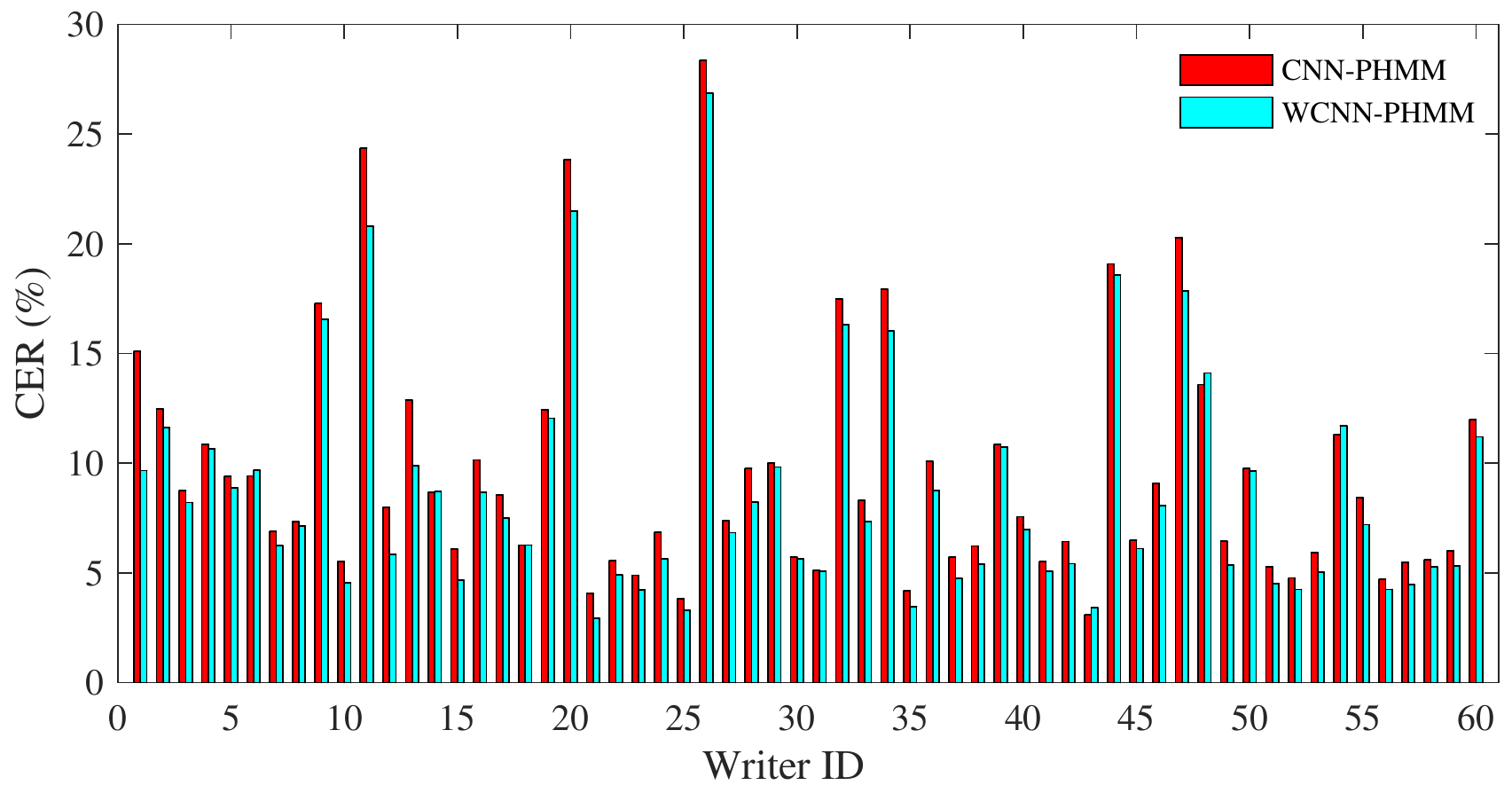}
\caption{CER (\%) comparison between WCNN-PHMM and CNN-PHMM for each writer of the competition set.}
\label{writers_error}
\end{figure}

To further demonstrate the effectiveness of writer adaptive training, we make a CER comparison between WCNN-PHMM and CNN-PHMM for each writer in Fig.~\ref{writers_error}. Consistent improvements could be obtained for most of the 60 writers, and there were only 5 exceptions (No. 6, No. 14, No. 43, No. 48, No. 54). Especially for those writers with relatively high CERs, significant gains could be achieved, e.g., the CER was reduced from 15.11\% to 9.66\% for writer No. 1, with a relative CER reduction of 36.1\%.

\begin{table}

\caption{CER (\%) comparison of different settings of adaptation layer number $P$ and writer code dimension $G$ in WCNN-PHMM.}
\centering \label{DiffLinks}
\scriptsize
\begin{tabular}{|c|c|c|c|c|c|c|c|c|c|}
\hline
$G$ & \multicolumn{7}{|c|}{200} & 100 & 400 \\
\hline
$P$ & 0 & 1 & 2 & 3 & 4 & 5 & 6 & 5 & 5 \\
\hline
CER & 9.54 & 9.29 & 9.17 & 9.04 & 8.99 & 8.96 & 8.96 & 9.05 & 9.02 \\
%
\hline
\end{tabular}
\end{table}

\subsubsection{WCNN with/without state tying}
\begin{figure}
\centering
\includegraphics[width=5in]{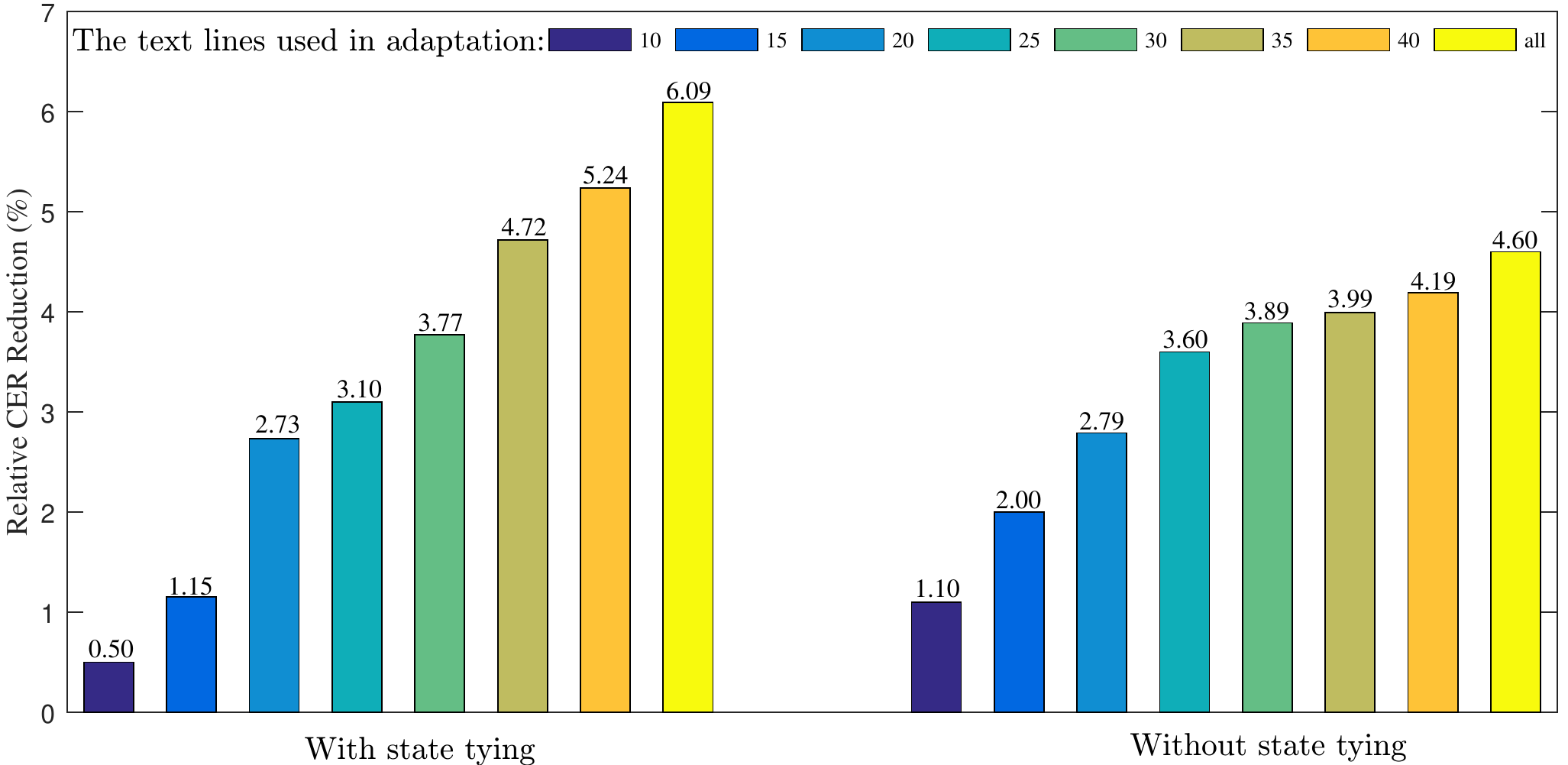}
\caption{The relative CER reduction (\%) of WCNN over CNN with/without state tying for different settings of text lines on the competition set.}
\label{influ_state_tying}
\end{figure}

In section~\ref{conf_wcnn}, we illustrated that WCNN could yield additional gains over CNN on top of PHMM using state tying. In this section, as shown in Fig.~\ref{influ_state_tying}, we compare the relative CER reduction (\%) in WCNN over CNN with/without state tying for different settings of text lines on the competition set.
For the CNN-HMM system without state tying, the best configured 5-state HMM in Table~\ref{DiffStates_CER} was used. In the competition set, the number of text lines for each writer ranged from 44 to 82. Overall, using all handwritten text lines of one writer for unsupervised adaptation, the CERs could be reduced from 10.02\% to 9.55\% (CNN-HMM vs. WCNN-HMM) and from 9.54\% to 8.96\% (CNN-PHMM vs. WCNN-PHMM). Those stable performance gains indicated that the proposed writer-adaptive training method was effective for systems with/without state tying (PHMM/HMM). Regarding the performance with respect to the amount of adaptation data, we observed that only 15 handwritten text lines for each writer on average could start to improve the recognition accuracy for unsupervised adaptation. When the number of text lines was reduced to 10, the relative CER reduction was limited, i.e., 0.5\% and 1.1\% for WCNN-PHMM and WCNN-HMM, respectively. Furthermore, when we continued to reduce the number of text lines to 5, the CERs increased compared with respective baselines. More interestingly, with increased adaptation data, the CER reduction in WCNN over CNN for the PHMM system with state tying became more significant than that for the HMM system without state tying, which implies that, as more handwritten data are collected from one writer, the proposed unsupervised adaptation via WCNN-PHMM can recognize handwritten text lines from this writer with more accuracy. Thus, the proposed WCNN-PHMM is a perfect demonstration of a compact model with adaptive capability.

\subsubsection{Multiple-pass decoding of WCNN-PHMM}
The basic intuition in the adaptation stage is better targets can promote the learning of the writer code and so produce beneficial feedback on the decoding results. By using the results of second-pass decoding based on WCNN-PHMM to generate better targets for the learning of the test writer codes, a third-pass decoding is conducted to get our final results. As shown in Table~\ref{Multi-pass}, the multiple-pass decoding can improve the recognition results (from 8.96\% to 8.64\%), which demonstrates that our intuition is right. We also list the run time comparison for different pass numbers. In order to make a fair comparison, all experiments here were evaluated on the same machine and we normalized the decoding time of first-pass to 1. The relative time consumption of $n$-pass ($n$=2,3) included two parts: the adaptation time and the decoding time. Although we could obtain a remarkable improvement via adaptation, the time consumption was linearly increased with the number of decoding passes. To address this problem, the acceleration of CNN and fast adaptation will be investigated in our future work.


\begin{table*}
\caption{CER (\%) and time consumption comparisons of multiple-pass decoding of WCNN-PHMM system.}
\centering \label{Multi-pass}
\begin{tabular}{|c|c|c|c|}
\hline
Multiple-pass Decoding            & CER (\%)   & Decoding Time & Adaptation Time  \\
\hline
First-pass (CNN-PHMM)             &  9.54      & 1.00             & 0.00   \\
\hline
Second-pass                       &   8.96     &    1.98           &  0.47  \\
\hline
Third-pass                        & \bf{8.64}  &      2.95         &  0.93  \\
\hline
\end{tabular}
\end{table*}


\subsubsection{Visualization analysis for writer code}
\begin{figure}
\centering
\subfigure[The t-SNE visualization of several writer codes.]{
\label{fig:subfig:a} 
\includegraphics[width=3in]{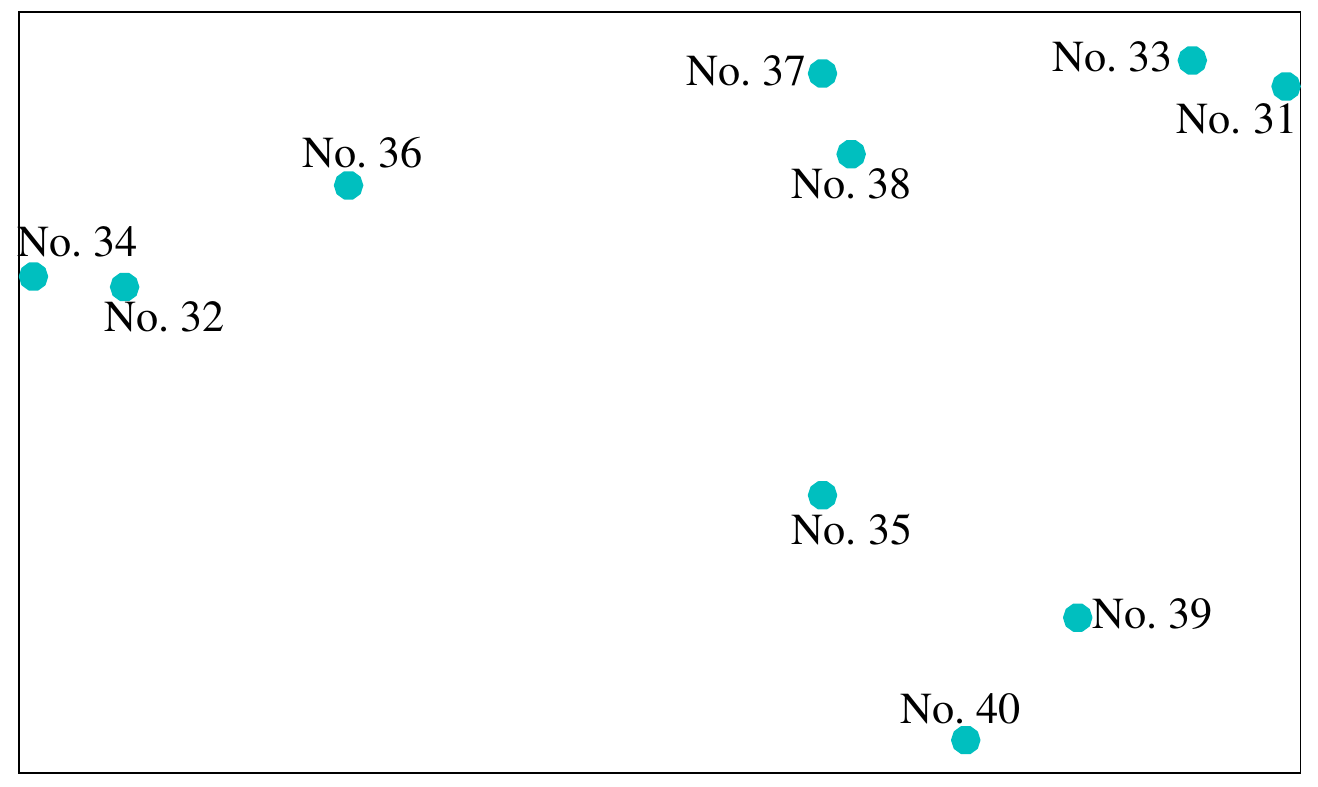}}
\subfigure[The corresponding handwriting examples of different writers.]{
\label{fig:subfig:b} 
\includegraphics[width=3in]{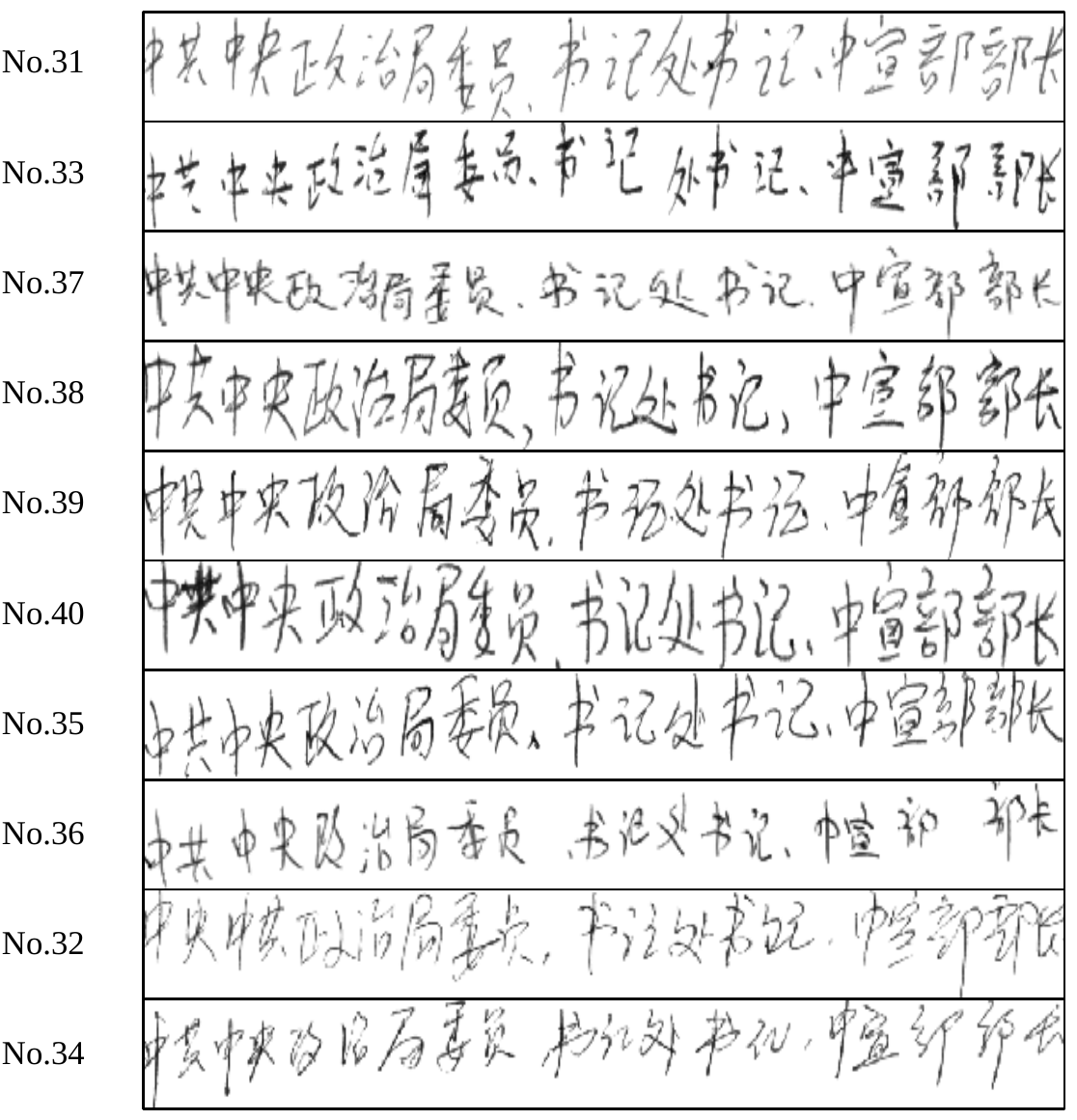}}
\caption{ Visualization analysis of several writer codes on the competition set. }
\end{figure}

To better understand why adaptation based on the writer code improves recognition performance, we adopted the t-SNE \cite{t-SNE} technique to visualize the generated writer codes by reducing its
dimension to 2. In Fig.~\ref{fig:subfig:a}, the distribution of several writer codes with the same transcripts on the competition set is shown. Correspondingly, we list their handwriting in Fig.~\ref{fig:subfig:b}. Interestingly, the distance between different writers in Fig.~\ref{fig:subfig:a} was a strong indicator of the similarity of the writing styles of different writers. For example, all the distances of ID pairs (31, 33), (32, 34), and (39, 40) were small, while the corresponding writing styles for those pairs were quite similar, as observed from the handwritten text lines, which demonstrates that the learned writer code indeed carries the writer information.

\subsection{Comparison of different language models}
Table~\ref{diff_lm} shows CER comparison of different language models. First, to demonstrate the scalability of our approach, we also conducted the corresponding 7360-class vocabulary experiments for different HMM systems. Please note that all the classes and writer data in HWDB1.0-HWDB1.2 were used in the 7360-class experiments rather than the subset listed in Table~\ref{info} that includes 3980-class experiments. Thus, the output layer sizes of CNN in the CNN-HMM system and WCNN in the WCNN-PHMM system were 36800 and 22080 for the 7360-class experiments, respectively, as illustrated in Fig.~\ref{development_WCNN_PHMM}. Although the confusion among the 7360 classes is higher, the CER of the 7360-class CNN-HMM was slightly increased from 10.02\% to 10.1\%, thus demonstrating the robustness of the HMM system. A surprising observation was that the CER of the 7360-class CNN-PHMM was remarkably reduced from 9.54\% in the 3980-class CNN-PHMM to 9.17\%, which might be due to the larger amount of training data used for 7360-class being better utilized and shared among different classes (compared with the 3980-class case) due to the use of our state-tying algorithm. Correspondingly, the recognition performance of WCNN-PHMM was also improved from the 3980-class case to the 7360-class case, i.e, 8.60\%, 8.42\% for the second-pass decoding and the third-pass decoding, respectively.

Second, by adding a language model, a great improvement could be obtained for all the systems. Besides, compared with the NLM, all systems that use the HLM performed better, e.g, a relative CER reduction of 6.3\%, 4.8\% and 4.8\% could be obtained in the 7000-class CNN-HMM, CNN-PHMM and WCNN-PHMM, respectively. It is reasonable that a weak character model could benefit more from a powerful language model.

\begin{table}
\caption{CER (\%) comparison of different language models.}
\centering \label{diff_lm}
\scriptsize
\begin{tabular}{|c|c|c|c|c|}
\hline
       Method   &  Vocabulary                & Without LM                & NLM          & HLM   \\
\hline
 \multirow{2}{*}{CNN-HMM}          &  3980         &   10.02    &   3.72           &      3.54       \\
 \cline{2-5}
                                   &  7360      &   10.1        &   3.82           &      3.58        \\
\hline
\multirow{2}{*}{CNN-PHMM}                   &  3980        &   9.54      &    3.57          &      3.44       \\
 \cline{2-5}
                                            &  7360       &   9.17       &    3.52          &      3.35      \\
\hline
\multirow{2}{*}{WCNN-PHMM }                  &  3980        &   8.64     &    3.39          &      3.27         \\
 \cline{2-5}                                  &  7360         &   8.42   &    3.33          &      3.17    \\
\hline
\end{tabular}
\end{table}

\subsection{Overall comparison and error analysis}



Table~\ref{all_results} shows an overall comparison of our proposed method and other state-of-the-art methods without/with a language model on the ICDAR 2013 competition set.
we list the state-of-the-art oversegmentation method heterogeneous CNN \cite{Wang16}, CNNs-RNNLM \cite{Yi17} and the segmentation-free method SMDLSTM-CTC \cite{Liu17ICDAR}, CNN-ACE \cite{JinCVPR19} in Table~\ref{all_results} for comparison. With the same configuration of vocabulary size (4 more garbage classes adopted in our HMM system), the proposed WCNN-PHMM yielded the best performance whether a language model was employed or not. Moreover, as shown in Table~\ref{diff_lm}, by using a powerful language model (HLM), the CNN-HMM, CNN-PHMM with one-pass decoding still could outperform the other methods.



\begin{table}
\caption{Performance comparison of our proposed method and other state-of-the-arts methods without/with language models on the 2013 ICDAR competition set.}
\centering \label{all_results}
\scriptsize
\begin{tabular}{|c|c|c|c|c|}
\hline
       Method   &  Vocabulary                 & Without LM    & With LM   \\
\hline
\multirow{2}{*}{WCNN-PHMM }                  &  3980        &   8.64          &        3.27         \\
 \cline{2-4}
                                             &  7360         &   8.42 &    3.17    \\
\hline \hline
        Wu \emph{et al.} \cite{Liu17ICDAR}    &  2672       &    9.98          &   7.39     \\
\cline{2-4}
                                                          &  7356       &    13.36          &    9.62    \\
\hline
        Wang \emph{et al.}  \cite{Wang16}    &   7356         &     11.21                      &    5.98    \\
\hline
      Wu \emph{et al.} \cite{Yi17}    &  7356    &  -  & 3.80 \\
\hline
      Xie \emph{et al.} \cite{JinCVPR19} & 7357    & 8.75 & 3.78 \\
\hline
\end{tabular}
\end{table}

\begin{figure}
\centering
\includegraphics[width=5in]{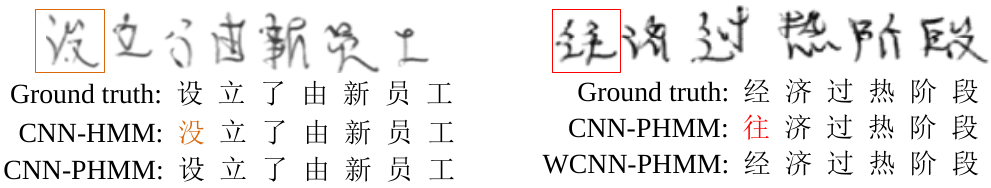}
\caption{Two examples of recognition results for different HMM systems.}
\label{error_analysis}
\end{figure}

\begin{figure}
\centering
\includegraphics[width=6.5in]{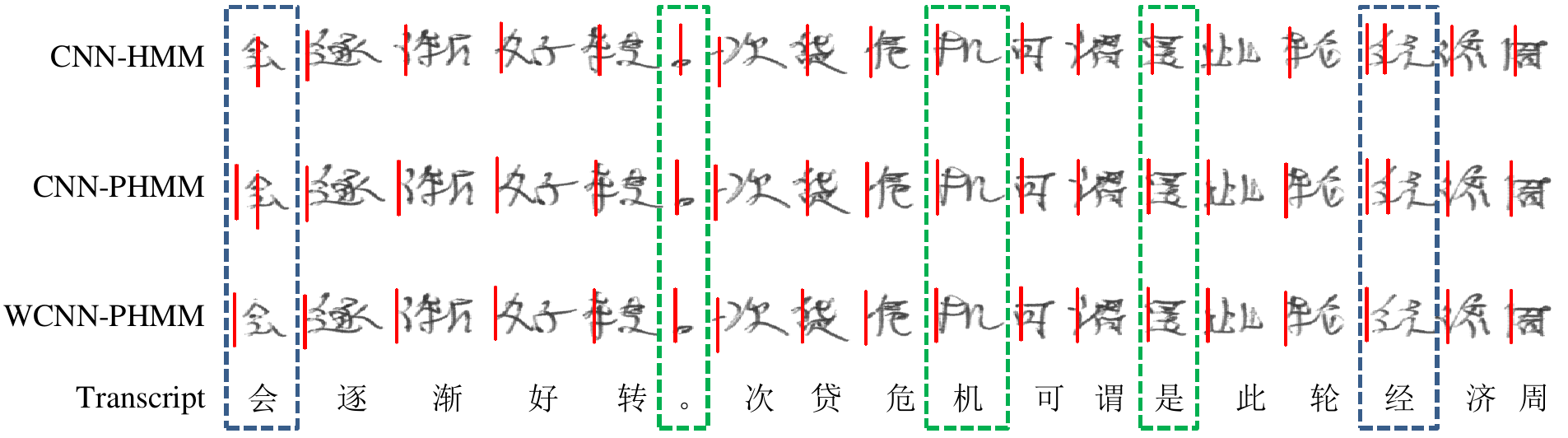}
\caption{Comparison of segmentation results of different HMM systems.}
\label{seg_info}
\end{figure}

For error analysis, we provide two examples in Fig.~\ref{error_analysis}. In the left part of the figure, the conventional CNN-HMM misrecognized the first character of the text line, while CNN-PHMM generated the correct result. A reasonable explanation is that the left radical of the character in the brown box became easier to recognized because state tying could potentially learn the parameters better than the radical with more shared training samples from other characters. In the right of the figure, CNN-PHMM made a substitution error (red), while WCNN-PHMM could correct this mistake. Arguably, even humans could confuse this handwritten character in isolation without any prior knowledge. However, by learning the writing style of this particular writer using the writer code, our WCNN-PHMM could correctly recognize it.
Besides, the HMM-based approaches can assign each image frame to a certain state belonging to a character. Once the process of recognition is completed, the segmentation information between different characters can be naturally found. Fig.~\ref{seg_info} shows the segmentation results of different HMM-based systems, i.e. CNN-HMM, CNN-PHMM and WCNN-PHMM. The red lines were the boundaries of different characters. For many characters such as the characters within the green dotted boxes, the CNN-PHMM and WCNN-PHMM provided more accurate boundaries than the CNN-HMM. For characters within the blue dotted boxes, we observed that the WCNN-PHMM could still give the right boundaries while the CNN-PHMM and CNN-HMM failed.

\begin{figure}
\centering
\subfigure[Comparison of reference state posterior probability of the frames for CNN-HMM and CNN-PHMM.]{
\label{fig:subfig:c} 
\includegraphics[width=3in]{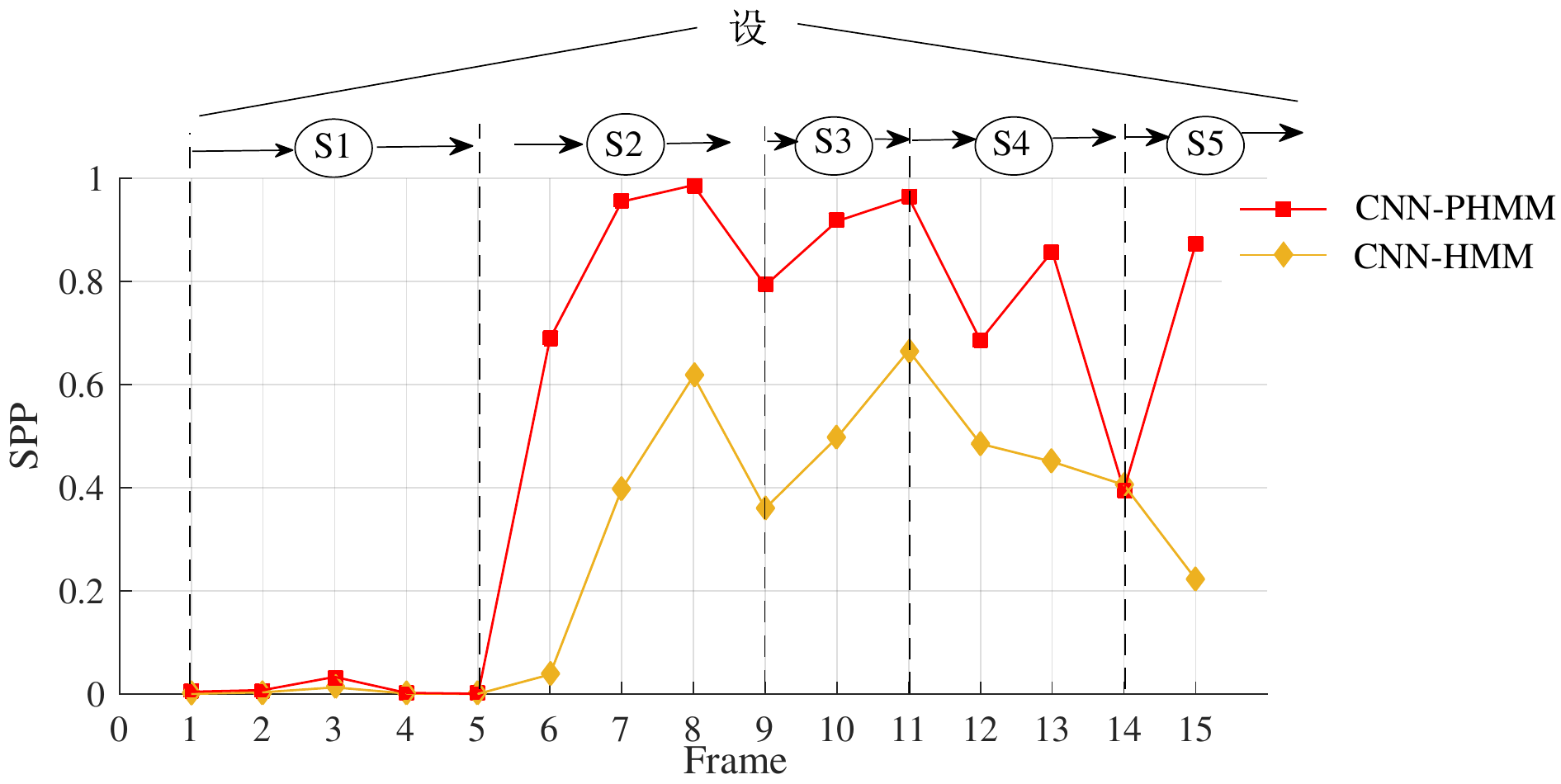}}
\subfigure[Comparison of reference state posterior probability (SPP) of the frames for CNN-PHMM and WCNN-PHMM.]{
\label{fig:subfig:d} 
\includegraphics[width=3in]{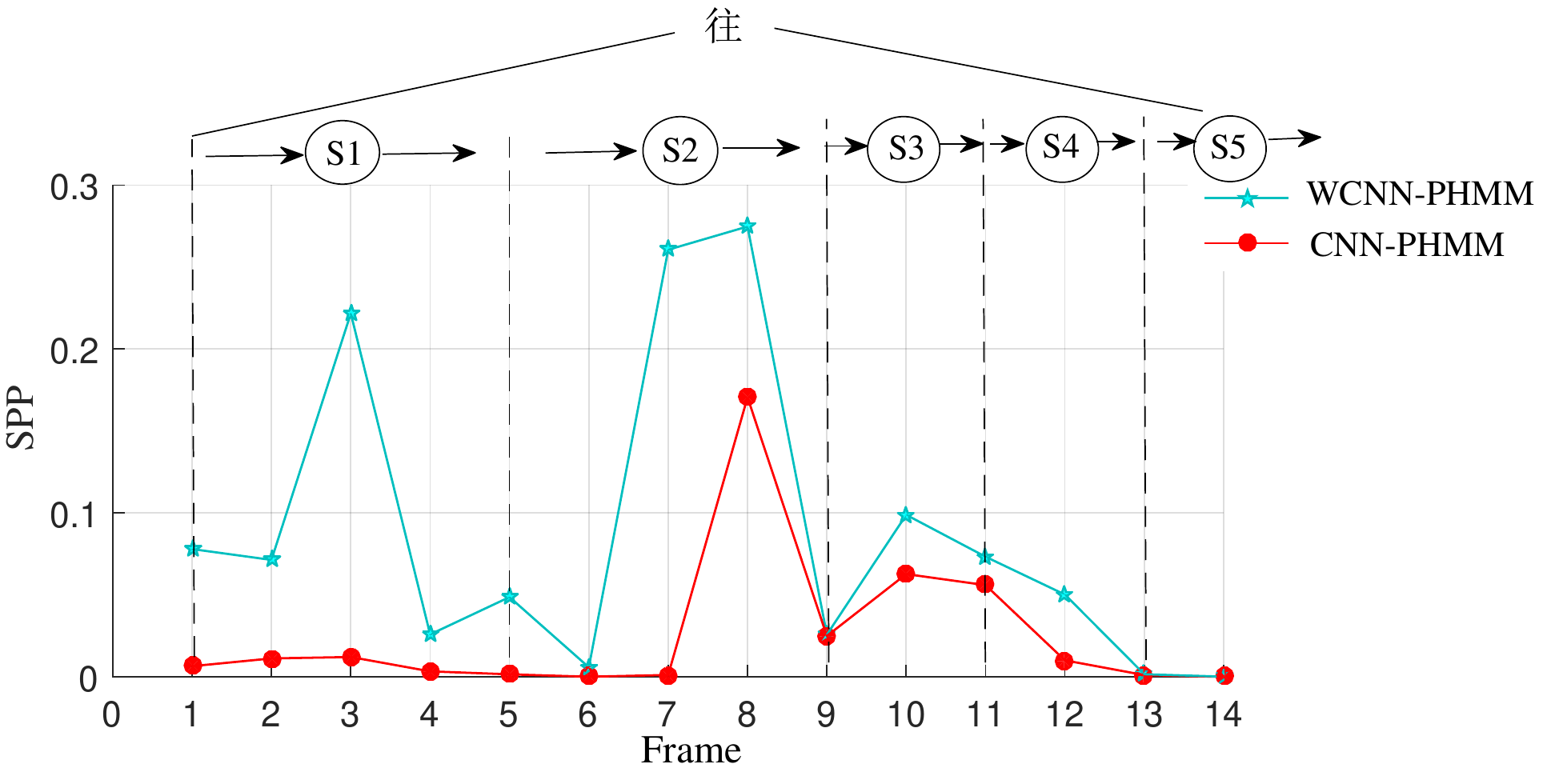}}
\caption{Comparison of reference state posterior probability (SPP) for different HMM systems. }
\end{figure}

Finally, in Figs.~\ref{fig:subfig:c} and~\ref{fig:subfig:d}, we explain and analyze the scores of the reference states of the underlying characters from the CNN outputs for CNN-HMM, CNN-PHMM, and WCNN-PHMM. Fig.~\ref{fig:subfig:c} shows the comparison of the state posterior probability (SPP) of the frames for the reference character class in the brown box of Fig.~\ref{error_analysis}. CNN-PHMM consistently generated higher SPPs than CNN-HMM for all frames of the sequence.  Similarly, in Fig.~\ref{fig:subfig:d}, corresponding to the character class in the red box of Fig.~\ref{error_analysis}, WCNN-PHMM always yielded higher SPPs than CNN-PHMM.



\section{Conclusion}
In this study, we propose a novel WCNN-PHMM architecture for offline handwritten Chinese text recognition to handle two key issues: the large vocabulary of Chinese characters and the diversity of writing styles. By combining parsimonious HMM based on state tying and unsupervised adaptation based on writer code, our new approach demonstrates its superiority to other state-of-the-art approaches according to both experimental results and analysis. However, current code-based adaptation simply depends on the backpropagation of network, which means adequate data is important. Besides, the 1-D HMM can not provide up-and-down information of characters. For future work, we will investigate the meta-learning to reduce dependence on data in adaptation and a more advanced way by using 2D-HMM to achieve recognition and segmentation. Furthermore, we will aim to accelerate the CNN to reduce decoding time.
\label{sec:con}

\section*{Acknowledgments}
This work was supported in part by the National Key R\&D Program of China under contract No. 2017YFB1002202, the National Natural Science Foundation of China under Grant Nos. 61671422 and U1613211, the Key Science and Technology Project of Anhui Province under Grant No. 17030901005, and the MOE-Microsoft Key Laboratory of USTC.

\ifCLASSOPTIONcaptionsoff
  \newpage
\fi

\end{document}